%% file: main.tex
\PassOptionsToPackage{dvipsnames,table}{xcolor}
\documentclass[10pt,twocolumn,letterpaper]{article}

\input{setup/package}
\input{setup/macros}
\input{setup/symbols}

\input{setup/graphicspath}
\usepackage[accsupp]{axessibility}

\makeatletter

\def\@fnsymbol#1{\ensuremath{\ifcase#1\or \dagger\or \ddagger\or
\mathsection\or \mathparagraph\or \|\or **\or \dagger\dagger
\or \ddagger\ddagger \else\@ctrerr\fi}}
\makeatother 

\graphicspath{{figures/}}

\usepackage{cvpr}              % To produce the CAMERA-READY version

\definecolor{cvprblue}{rgb}{0.21,0.49,0.74}
\usepackage[pagebackref,breaklinks,colorlinks,linkcolor=red,citecolor=cvprblue,urlcolor=blue,bookmarks=false]{hyperref}

\def\paperID{4583} % *** Enter the CVPR Paper ID here
\def\confName{CVPR}
\def\confYear{2024}

\begin{document}

\title{
In-N-Out: Faithful 3D GAN Inversion with Volumetric Decomposition for Face Editing
}

\author{Yiran Xu $^{1,2}$ \hspace{0.75cm} Zhixin Shu$^{2}$ \hspace{0.75cm} Cameron Smith$^{2}$ \hspace{0.75cm} Seoung Wug Oh$^{2}$ \hspace{0.75cm}  Jia-Bin Huang$^{1}$ \\
\vspace{-0.2cm}\\
$^{1}$University of Maryland, College Park, $^{2}$Adobe Research\\
% \vspace{-0.2cm}\\
\url{https://in-n-out-3d.github.io/}
}

\twocolumn[{
    \renewcommand\twocolumn[1][]{#1}
    \maketitle
    \vspace{-12mm}
    \input{figures/teaser.tex}    }]
\maketitle
\thispagestyle{empty}
\input{0_abstract}

\input{1_introduction}
\input{2_related}
\input{3_method}
\input{4_result}
\input{5_limitations}
\input{6_conclusions}

%%%%%%%%% REFERENCES
{\small
\bibliographystyle{ieee_fullname}
\bibliography{main}
}

\end{document}

% \onecolumn
% {
% \Large
% \begin{center}
% \textbf{\noindent 3D Photography using Context-aware Layered Depth Inpainting\\Supplementary Material}
% \end{center}
% \maketitle
% }

\title{
\large{In-N-Out: Faithful 3D GAN Inversion with Volumetric Decomposition for Face Editing}
\\ \large{Supplementary Material} 
% \\ \large{Paper ID: 4583}
}

% \author{
%       XXX XXX \textsuperscript{\dag}
% \quad XXX \textsuperscript{\dag}
% \quad XXX \textsuperscript{\S}
% \vspace{0.5em}\\
% \textsuperscript{\dag} XXX University
% \qquad \textsuperscript{\S} XXX Company
% \vspace{0.2em}\\
% {\tt\small \textsuperscript{\dag}\{xxx\}@xxx.edu}, \quad
% {\tt\small \textsuperscript{\S}xxx@xxx.edu}
% }

% \twocolumn[{
% \renewcommand\twocolumn[1][]{#1}
% \maketitle
% }]
\maketitle

% Main paper
\input{supp_content}

\clearpage
{\small
\bibliographystyle{ieee_fullname}
\bibliography{main}
}

%% file: setup/package.tex
\usepackage{subcaption}
\usepackage{float}
\usepackage[justification=raggedright]{caption}	% makes captions ragged right - thanks to Bryce Lobdell
\usepackage{lscape}                                         % Useful for wide tables or figures.
\usepackage{wrapfig}

\usepackage[lined,ruled,linesnumbered]{algorithm2e}
\usepackage{animate} %% convert frames to video

\usepackage{booktabs}                   % Publication quality tables
\usepackage{multirow}

\usepackage{makecell}

\usepackage{paralist}
\usepackage{enumitem}
\setlist[enumerate]{itemsep=0mm}

\usepackage{bm}                          % Make bold, italic math symbols
\usepackage{epsfig}                      % for figures
\usepackage{graphicx}                  % another package that works for figures
\usepackage{times}
\usepackage{mathtools}
\usepackage{amssymb,amsmath}   % Short math guide for LaTeX ftp://ftp.ams.org/pub/tex/doc/amsmath/short-math-guide.pdf

\usepackage{units}
\usepackage{color}

\usepackage{comment}

\RequirePackage[hyphens]{url}
\definecolor{cvprblue}{rgb}{0.21,0.49,0.74}
\usepackage[pagebackref,breaklinks,colorlinks,linkcolor=red,citecolor=cvprblue,urlcolor=blue,bookmarks=false]{hyperref}
\usepackage{xspace}
\usepackage[table]{xcolor}
\usepackage{setspace}
\usepackage{grfext}
\PrependGraphicsExtensions*{.jpg,.png,.PNG}
\usepackage{pifont}
\usepackage{multirow}
\usepackage{bigstrut}
\usepackage{tabularx}

%% file: setup/macros.tex
\usepackage{soul}
\usepackage{svg}

\usepackage[accsupp]{axessibility} % Improves PDF readability for those with disabilities.

\def\eg{e.g.,~}               % for example
\def\ie{i.e.,~}               % that is, in other words
\DeclareMathOperator*{\argmin}{\arg\!\min}

\DeclareMathAlphabet{\altmathcal}{OMS}{cmsy}{m}{n}
\DeclareMathAlphabet{\mathbfit}{OT1}{ptm}{bx}{it}

\newlength\paramargin
\newlength\figmargin

\newlength\secmargin
\newlength\figcapmargin
\newlength\tabcapmargin

\newcommand{\mpage}[2]
{
\begin{minipage}{#1\linewidth}\centering
#2
\end{minipage}
}

\newcommand{\topic}[1]
{
\vspace{1mm}\noindent\textbf{#1}
}

\long\def\ignorethis#1{}
\newbox\jsavebox%

\makeatletter
\newcommand{\providelength}[1]{%
  \@ifundefined{\expandafter\@gobble\string#1}
   {% if #1 is undefined, do \newlength
    \typeout{\string\providelength: making new length \string#1}%
    \newlength{#1}%
   }
   {% else check whether #1 is actually a length
    \sdaau@checkforlength{#1}%
   }%
}

\newcommand{\sdaau@checkforlength}[1]{%
  \edef\sdaau@temp{\expandafter\sdaau@getfive\meaning#1TTTTT$}%
  \ifx\sdaau@temp\sdaau@skipstring
    \typeout{\string\providelength: \string#1 already a length}%
  \else
    \@latex@error
      {\string#1 illegal in \string\providelength}
      {\string#1 is defined, but not with \string\newlength}%
  \fi
}
\def\sdaau@getfive#1#2#3#4#5#6${#1#2#3#4#5}
\edef\sdaau@skipstring{\string\skip}
\makeatother
\usepackage[capitalize]{cleveref}
\crefname{section}{Sec.}{Secs.}
\Crefname{section}{Section}{Sections}
\Crefname{table}{Table}{Tables}
\crefname{table}{Tab.}{Tabs.}

%% file: setup/symbols.tex
\def\xi{\mathbf{x}_i}

%% file: setup/graphicspath.tex
\graphicspath{{figures}, {examples}}

%% file: figures/teaser.tex
% \begin{figure*}
\begin{center}
\centering

\hspace{-5.5mm}
$
\mpage{0.01}{\raisebox{3pt}{\rotatebox{90}{Single Image Input}}}\left\{ \begin{array}{rrrrrr}
     {\frame{\includegraphics[trim=0 0 0 0, clip,width=.145\textwidth]{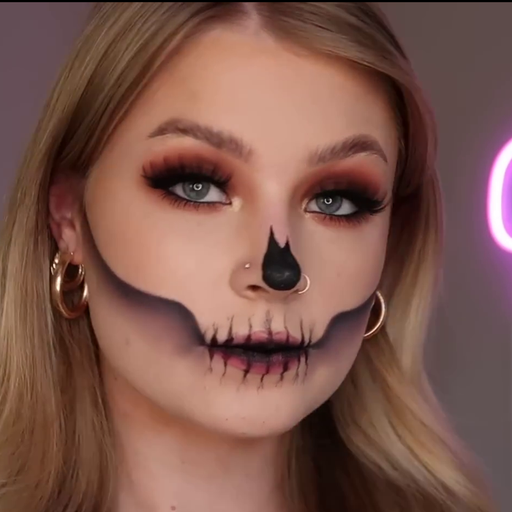}}}\hfill\hspace{2mm} & {\frame{\includegraphics[trim=0 0 0 0, clip,width=.145\textwidth]{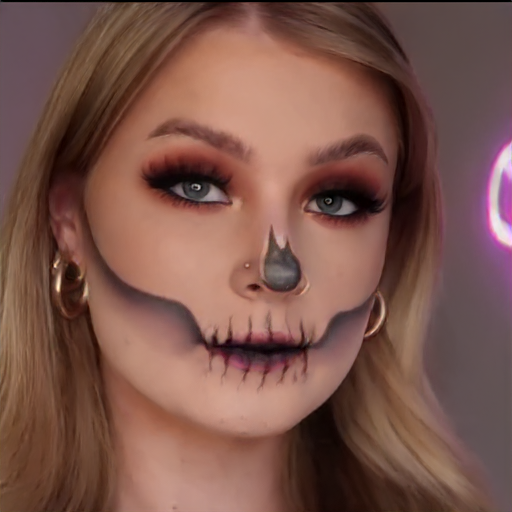}}}\hfill & \frame{\includegraphics[trim=0 0 0 0, clip,width=.145\textwidth]{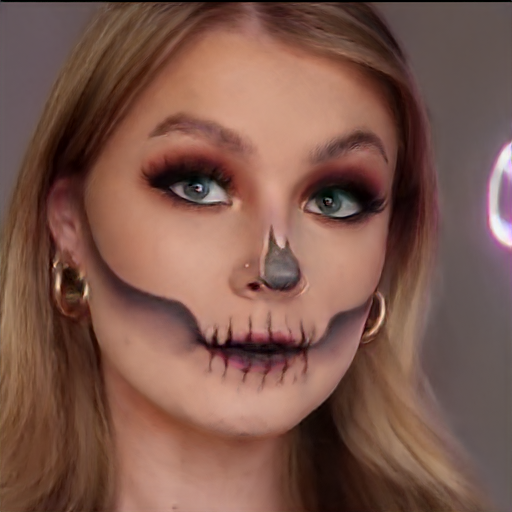}}\hfill & {\frame{\includegraphics[trim=0 0 0 0, clip,width=.145\textwidth]{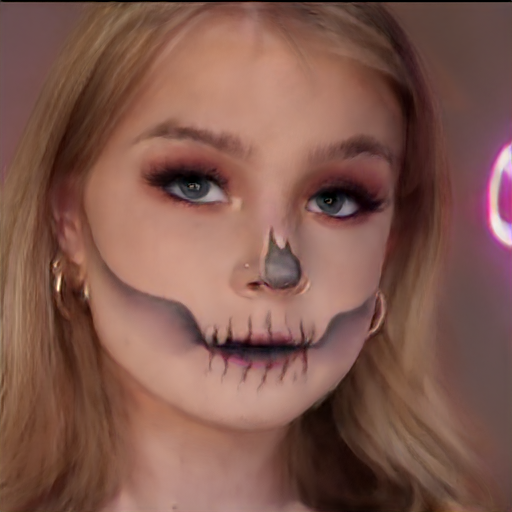}}}\hfill 
     & {\frame{\includegraphics[trim=0 0 0 0, clip,width=.145\textwidth]{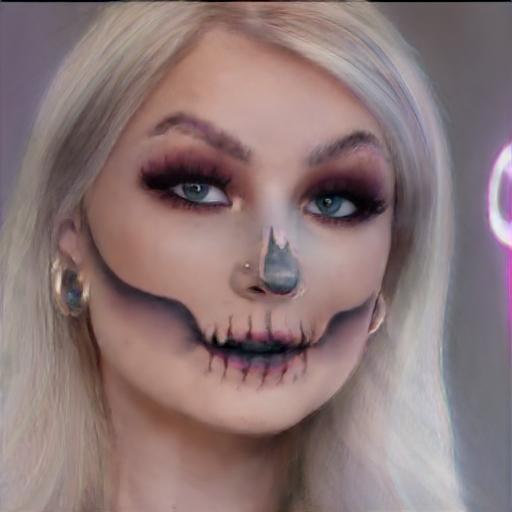}}}\hfill
     & {\frame{\includegraphics[trim=0 0 0 0, clip,width=.145\textwidth]{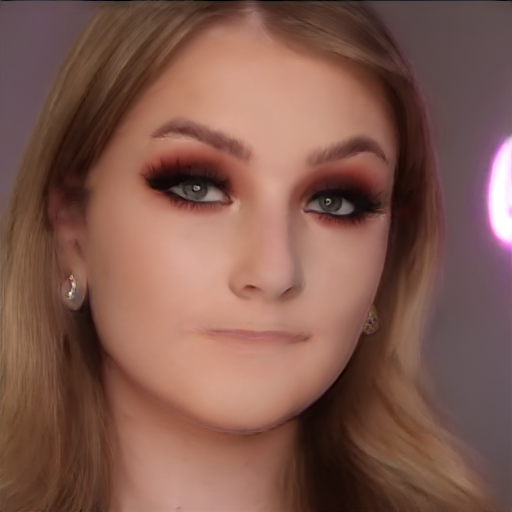}}}\\
     \vspace{-1mm}
     & \\
     % \text{Input}\hfill\hspace{2mm} & \text{Reconstruction}\hspace{2mm}\hfill &  \textit{Surprised}\hspace{5mm}\hfill & \textit{Younger}\hspace{6mm}\hfill & \textit{Elsa}\hspace{9mm}\hfill & \text{OOD removal}\hfill \\
     \frame{\includegraphics[trim=0 0 0 0, clip,width=.145\textwidth]{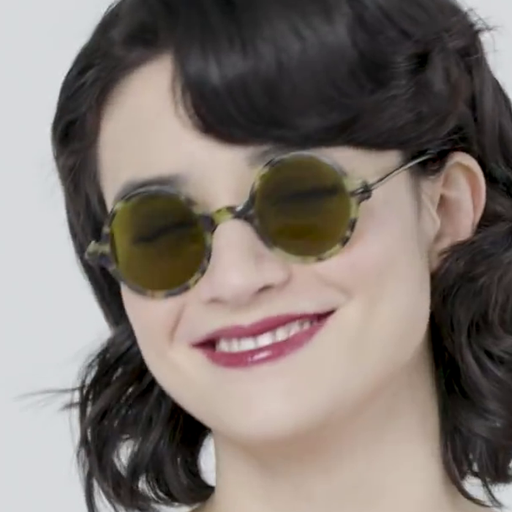}}\hfill\hspace{2mm}
     & \frame{\includegraphics[trim=0 0 0 0, clip,width=.145\textwidth]{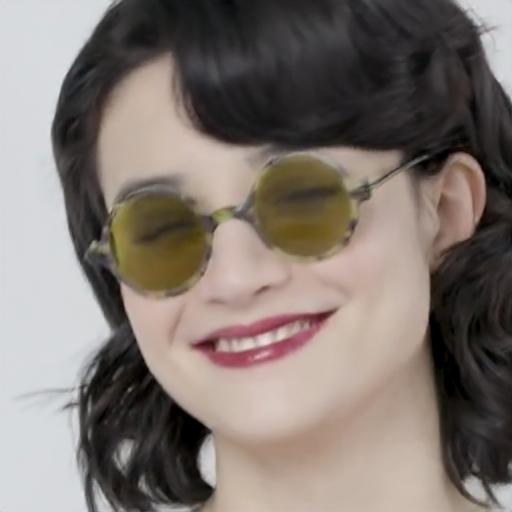}}\hfill 
     & {\frame{\includegraphics[trim=0 0 0 0, clip,width=.145\textwidth]{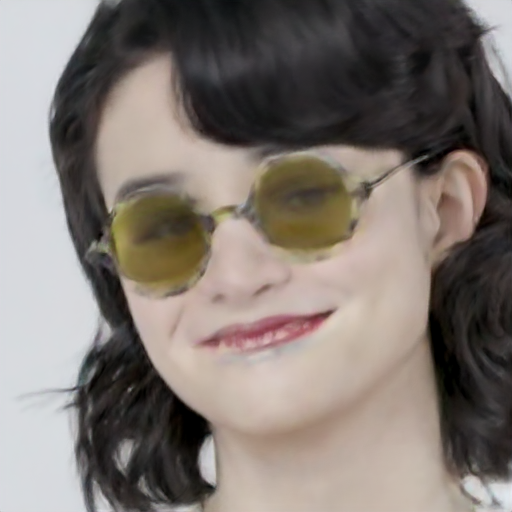}}}\hfill
     & \frame{\includegraphics[trim=0 0 0 0, clip,width=.145\textwidth]{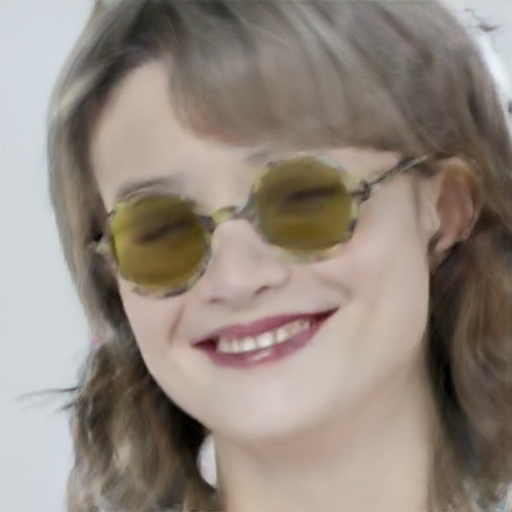}}\hfill 
     & \frame{\includegraphics[trim=0 0 0 0, clip,width=.145\textwidth]{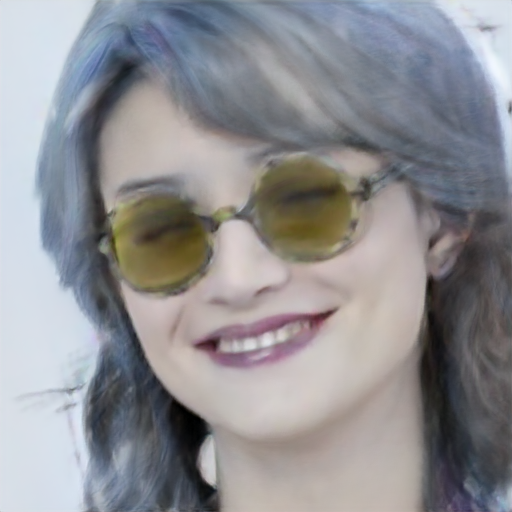}}\hfill
     & \frame{\includegraphics[trim=0 0 0 0, clip,width=.145\textwidth]{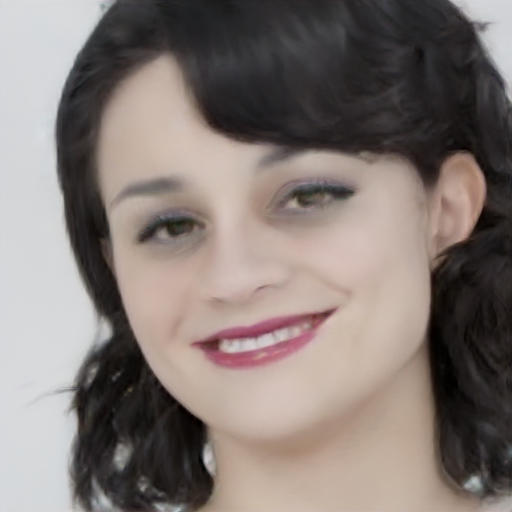}}\\
     % \text{Input}\hfill\hspace{2mm} & \text{Reconstruction}\hfill &  \textit{Less Smile}\hfill & \textit{Blond}\hfill & \textit{Elsa}\hfill & \text{OOD removal} \\
\end{array}
\right.
$

% \mpage{0.01}{\raisebox{3pt}{\rotatebox{90}{Single Image Input}}}
% {\frame{\includegraphics[trim=0 0 0 0, clip,width=.150\textwidth]{images/teaser/target_00000-teaser-Halloween1.png}}}\hfill\hspace{2mm}
% {\frame{\includegraphics[trim=0 0 0 0, clip,width=.150\textwidth]{images/teaser/proj_00000-inv-teaser-Halloween1.png}}}\hfill
% \frame{\includegraphics[trim=0 0 0 0, clip,width=.150\textwidth]{images/teaser/proj_00000-surprised-teaser-Halloween1.png}}\hfill
% {\frame{\includegraphics[trim=0 0 0 0, clip,width=.150\textwidth]{images/teaser/proj_00000-younger-teaser-Halloween1.png}}}\hfill
% {\frame{\includegraphics[trim=0 0 0 0, clip,width=.150\textwidth]{images/teaser/proj_00000-elsa-teaser-Halloween1.png}}}\hfill
% {\frame{\includegraphics[trim=0 0 0 0, clip,width=.150\textwidth]{images/teaser/proj_00000-teaser-removal-Halloween1.png}}}\\
\vspace{-30mm}
\mpage{0.17}{{Input}}\hfill
\mpage{0.15}{{Recon.}} \hfill
\mpage{0.15}{\textit{Surprised}} \hfill
\mpage{0.15}{\textit{Younger}} \hfill
\mpage{0.15}{\textit{Elsa}} \hfill
\mpage{0.15}{{OOD Removal}}

% \mpage{0.01}{\raisebox{95pt}{\rotatebox{90}{ }}} 
% \frame{\includegraphics[trim=0 0 0 0, clip,width=.150\textwidth]{images/teaser/target_00000-teaser-sunglasses1.png}}\hfill\hspace{2mm}
% \frame{\includegraphics[trim=0 0 0 0, clip,width=.150\textwidth]{images/teaser/proj_00000-teaser-inv-sunglasses1.png}}\hfill
% {\frame{\includegraphics[trim=0 0 0 0, clip,width=.150\textwidth]{images/teaser/proj_00000-teaser-smile--sunglasses1.png}}}\hfill
% \frame{\includegraphics[trim=0 0 0 0, clip,width=.150\textwidth]{images/teaser/proj_00000-teaser-blond-sunglasses1.png}}\hfill
% \frame{\includegraphics[trim=0 0 0 0, clip,width=.150\textwidth]{images/teaser/proj_00000-teaser-elsa-sunglasses1.png}}\hfill
% \frame{\includegraphics[trim=0 0 0 0, clip,width=.150\textwidth]{images/teaser/proj_00000-teaser-removal-sunglasses1.png}}\\
\vspace{25mm}
\mpage{0.17}{{Input}}\hfill
\mpage{0.15}{{Recon.}} \hfill
\mpage{0.15}{\textit{Less smile}} \hfill
\mpage{0.15}{\textit{Blond}} \hfill
\mpage{0.15}{\textit{Elsa}} \hfill
\mpage{0.15}{{OOD Removal}} \\
\hspace{-6.5mm}
\mpage{0.01}{\raisebox{70pt}{\rotatebox{90}{}}} 
\mpage{0.01}{\raisebox{70pt}{\rotatebox{90}{Video input}}} 
\frame{\includegraphics[trim=0 0 0 0, clip,width=.150\textwidth]{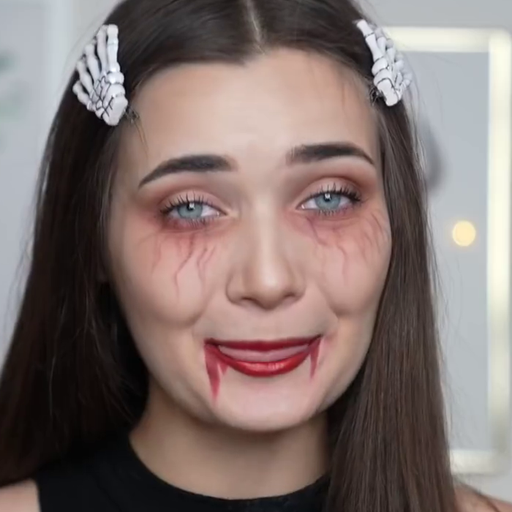}}\hspace{-22mm}\hfill
\frame{\includegraphics[trim=0 0 0 0, clip,width=.150\textwidth]{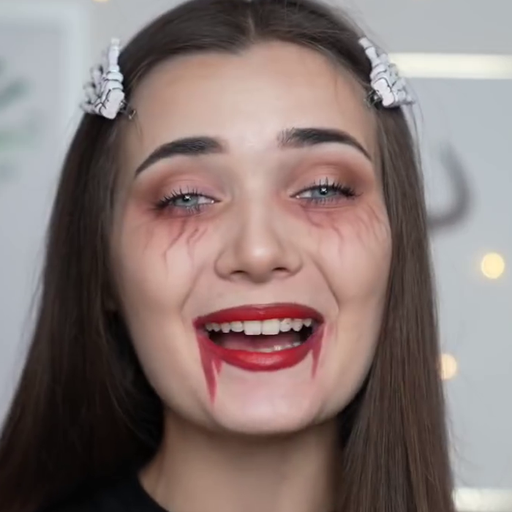}}\hspace{-22mm}\hfill
\frame{\includegraphics[trim=0 0 0 0, clip,width=.150\textwidth]{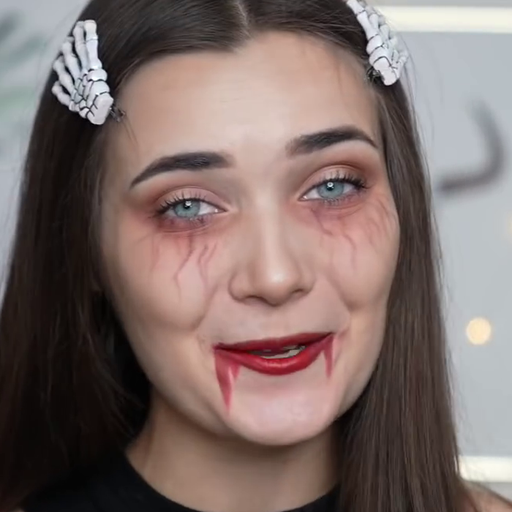}}\hspace{-20mm}\hfill
\hspace{-5mm}
\mpage{0.01}{\raisebox{70pt}{\rotatebox{90}{Video input}}}
\frame{\includegraphics[trim=0 0 0 0, clip,width=.150\textwidth]{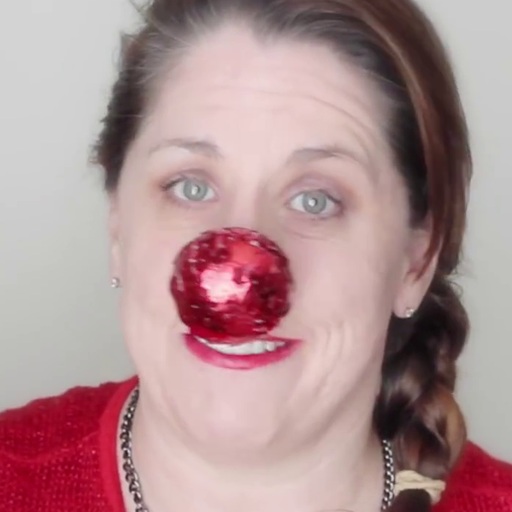}}\hspace{-22mm}\hfill
\frame{\includegraphics[trim=0 0 0 0, clip,width=.150\textwidth]{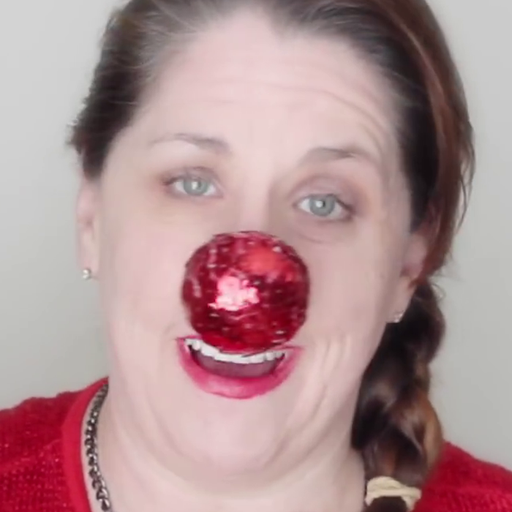}}\hspace{-22mm}\hfill
\frame{\includegraphics[trim=0 0 0 0, clip,width=.150\textwidth]{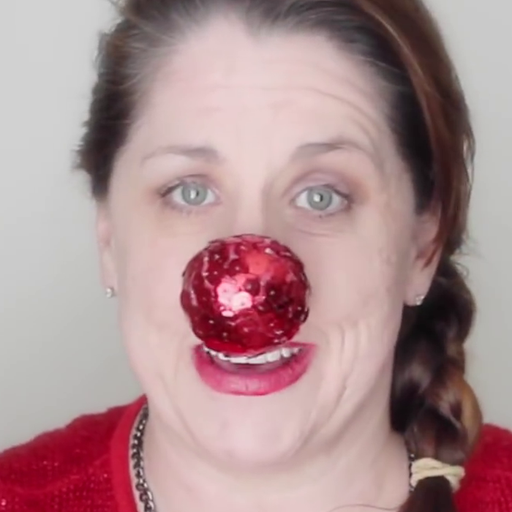}}\hspace{-22mm}\\
\vspace{-18mm}
\hspace{-6.5mm}
\mpage{0.01}{\raisebox{70pt}{\rotatebox{90}{}}} 
\mpage{0.01}{\raisebox{70pt}{\rotatebox{90}{\textit{Elsa}}}}
\frame{\includegraphics[trim=0 0 0 0, clip,width=.150\textwidth]{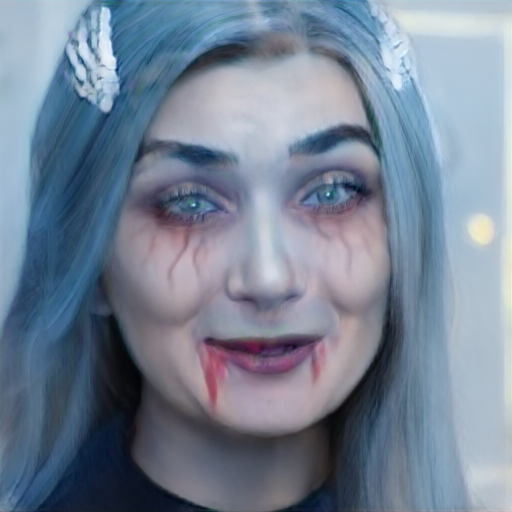}}\hspace{-22mm}\hfill
\frame{\includegraphics[trim=0 0 0 0, clip,width=.150\textwidth]{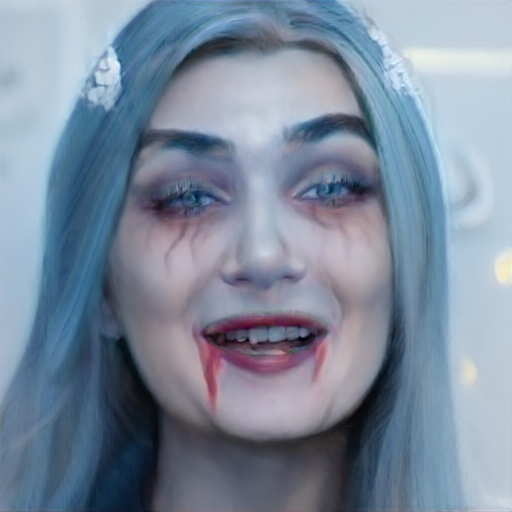}}\hspace{-22mm}\hfill
\frame{\includegraphics[trim=0 0 0 0, clip,width=.150\textwidth]{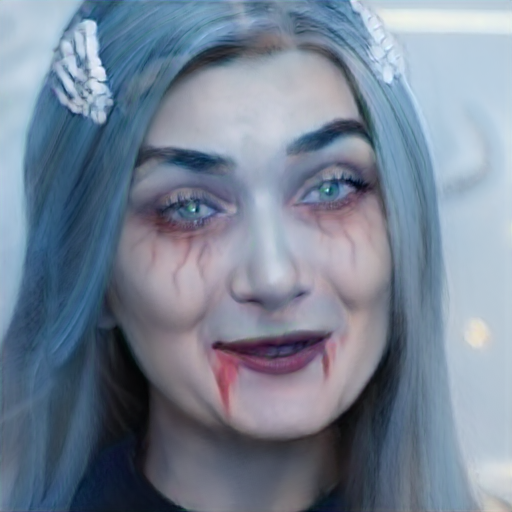}}\hspace{-20mm}\hfill
\hspace{-5mm}
\mpage{0.01}{\raisebox{70pt}{\rotatebox{90}{\textit{eyeglasses}}}}
\frame{\includegraphics[trim=0 0 0 0, clip,width=.150\textwidth]{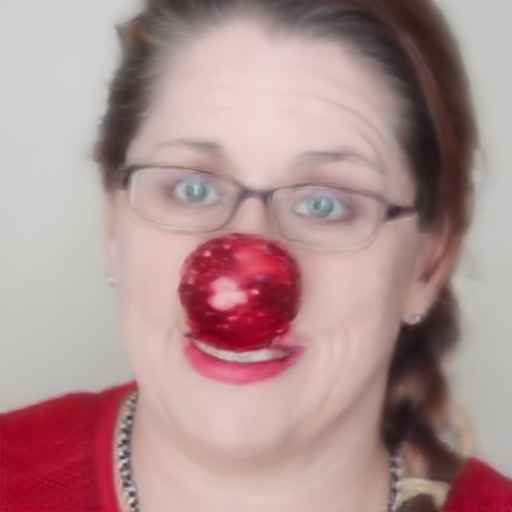}}\hspace{-22mm}\hfill
\frame{\includegraphics[trim=0 0 0 0, clip,width=.150\textwidth]{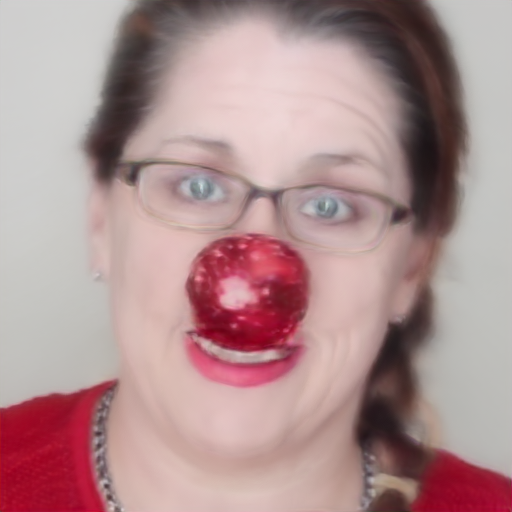}}\hspace{-22mm}\hfill
\frame{\includegraphics[trim=0 0 0 0, clip,width=.150\textwidth]{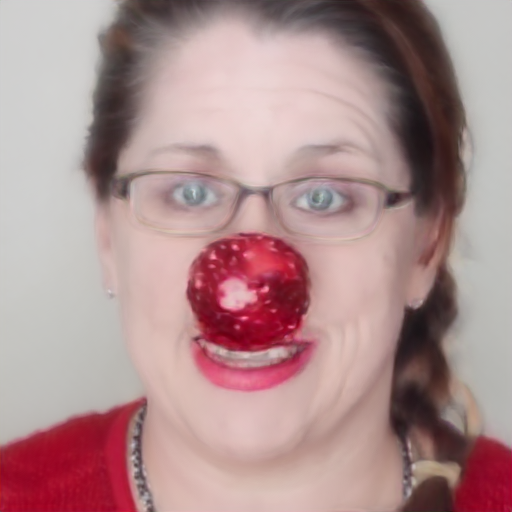}}\hspace{-22mm}\\

\vspace{-20mm}

\captionof{figure}{
\textbf{Semantic editing for \emph{out-of-distribution} data.} 
We present a method for reconstructing and editing an out-of-distribution (OOD) image or video using a pre-trained 3D-aware generative model (EG3D~\cite{chan2022efficient}). 
Our method explicitly models and reconstructs the occluders in 3D, allowing faithful reconstruction of the input while preserving the semantic editing capability. 
Here we showcase the reconstruction and editing results ``Less smile'', ``Younger'', ``Blond''~\cite{shen2020interpreting}, ``Elsa'', ``Surprised''~\cite{patashnik2021styleclip}.
Our method can also remove the OOD part.
Data are from the Internet (Creative Commons).
}
\label{fig:teaser}
\end{center}

%% file: 0_abstract.tex
\begin{abstract}
3D-aware GANs offer new capabilities for view synthesis while preserving the editing functionalities of their 2D counterparts.
%These methods use GAN inversion to reconstruct images or videos by optimizing a latent code, allowing for semantic editing by manipulating the code.
GAN inversion is a crucial step that seeks the latent code to reconstruct input images or videos, subsequently enabling diverse editing tasks through manipulation of this latent code. 
However, a model pre-trained on a particular dataset (\eg, FFHQ) often has difficulty reconstructing images with out-of-distribution (OOD) objects such as faces with heavy make-up or occluding objects. 
% What have we done
We address this issue by explicitly modeling OOD objects from the input in 3D-aware GANs.
% Core idea
Our core idea is to represent the image using two individual neural radiance fields: one for the in-distribution content and the other for the out-of-distribution object. The final reconstruction is achieved by optimizing the composition of these two radiance fields with carefully designed regularization.
We demonstrate that our explicit decomposition alleviates the inherent trade-off between reconstruction fidelity and editability.
% Evaluation
We evaluate reconstruction accuracy and editability of our method on challenging real face images and videos and showcase favorable results against other baselines.
More results can found at \href{https://in-n-out-3d.github.io/}{https://in-n-out-3d.github.io/}.
\end{abstract}

% The project page is available at \hyperlink{https://in-n-out-3d.github.io/}{https://in-n-out-3d.github.io/}. 

%% file: 1_introduction.tex
\section{Introduction}
\label{sec:intro}

% Storyline
% Problem: 3D GAN face (with OOD objects) inversion and editing (from single frame/multi-frames)
% Why challenging: only using in-distribution representation (tri-plane) is hard to reconstruct the OOD object or edit the face. 
% What previous people have done: Finetuning G (PTI, W+) -> reduce the editability, encoder-based methods -> bad reconstruction, using pseudo-view (HFGI3D, single view) -> unsatisfactory geometry and editing.
% Our idea: Using two tri-planes, one for in-distribution and one for OOD. 

% GANs are awesome
% \topic{GAN-based editing.} 
\textit{GAN inversion} ~\cite{xia2022gan,zhu2020domain,abdal2021styleflow,richardson2021encoding,tov2021designing} is a set of techniques that project an input image onto the latent space of a pre-trained GAN to obtain a latent code so that the image generator can reconstruct the input.
%(so that the input image can be reconstructed by feeding the latent code into a pre-trained GAN). 
%By changing the latent code, 
This is particularly useful as one could perform various creative semantic editing tasks~\cite{shen2020interpreting,harkonen2020ganspace,patashnik2021styleclip,gal2022stylegan} for images.
Similar techniques have also been applied in the video domain, with which recent methods also achieved temporally consistent editing~\cite{tzaban2022stitch,xu2022temporally}.
%However, these methods focus on 2D GANs and lack explicit control over the viewpoints. 
However, the majority of these methods are effective primarily with 2D GANs, and they fall short in offering explicit 3D controllability, such as view synthesis capabilities. 
% 3D GANs are awesome
% \topic{3D-aware GANs.} 
With the rapid recent advancements in 3D reconstruction, especially in neural radiance fields (NeRFs)~\cite{mildenhall2020nerf,barron2021mip,mueller2022instant,Chen2022ECCV}, high-quality 3D-aware GANs~\cite{gu2022stylenerf,orel2022styleSDF,chan2022efficient,skorokhodov2022epigraf} have emerged as a powerful tool for learning 3D generation from 2D images.
3D-aware GANs, equipped with a 3D representations like NeRFs~\cite{gu2022stylenerf,chan2022efficient} or SDF~\cite{orel2022styleSDF}, offer explicit control over camera views and ensure 3D geometric consistency in generation. Additionally, they retain the generative capacity and editability of 2D GANs~\cite{karras2019style,karras2020analyzing,karras2020training,karras2021alias}.
This enables applications such as novel view synthesis, semantic image editing~\cite{xie2023hfgi3d,Yuan_2023_GOAE,lan2023e3dge,yin2023spi,sun2022ide,Simsar_2023_swap3d} and video editing~\cite{fruhstuck2023vive3d,trevithick2023real}.

\topic{Core challenges.} 
While state-of-the-art 3D GAN inversion methods achieve remarkable advances in both image and video editing for human faces, 
they face challenges when dealing with images including \textit{out-of-distribution (OOD)} objects (\eg, heavy make-ups or occlusions). 
% their capacity to deal with a face with \textit{out-of-distribution (OOD)} objects (\eg, heavy make-ups or occlusions) is not satisfactory. 
This limitation arises primarily because these models are pre-trained only on natural faces without complex textures or substantial occlusions. 
%, absent of complicated texture or large occlusion, and
As a result, the editability performance deteriorates when a pre-trained GAN is forced to model OOD objects in the GAN inversion process.
This is commonly known as the \emph{reconstruction-editability trade-off}~\cite{tov2021designing}.
Existing GAN inversion methods assume that a \emph{single} latent code corresponding to the input image can be found in the latent space~\cite{vsubrtova2022chunkygan,xia2022gan} through optimization once the model is trained. Therefore, they aim to reconstruct the in-distribution (InD) content (e.g., natural face) and the OOD objects \emph{together}.
However, OOD components often cannot be well modeled in a pre-trained GAN, and consequently cannot be well represented with it using a single latent code, existing methods either cannot reconstruct them faithfully~\cite{sun2022ide} 
or can reconstruct them (e.g., through fine-tuning the generator) but alters the latent space properties and deteriorates the editability~\cite{roich2022pivotal} (Figure~\ref{fig:motivation}).
\input{figures/motivation.tex}

% What we have done? Start with ``In this paper, "
% - Provide forward references
% Our main idea in one word
\topic{Our work.} 
We propose a new approach to address this issue by drawing inspiration from recent composite volume rendering works that compose multiple radiance fields during
rendering~\cite{martin2021nerf,gao2021dynamic,yang2021learning,wu2022d}.
Our core idea is to 
\emph{decompose} the 3D representation of an image with OOD components into an \textit{in-distribution} (InD) part and an \textit{out-of-distribution} part, and compose them together to reconstruct the image in a composite volumetric rendering manner.
We use EG3D~\cite{chan2022efficient} as our 3D-aware GAN backbone and leverage its tri-plane representation to model this composed rendering pipeline. 
% Two radiance fields are used.
For the InD component (\ie natural face), we project pixel values onto EG3D's $\mathcal{W}{+}$ space for an InD component reconstruction. 
%For the OOD part, we use an additional tri-plane to represent it. 
We further introduce an additional tri-plane to represent the OOD content.
After that, we combine these two radiance fields in a composite volumetric rendering to reconstruct the input frames.
During the editing stage, we perform the latent code based editing solely on the InD part and leave the OOD component unaltered.
%, \ie the latent code $w$, \emph{without} touching the OOD part. 
%Any 
This framework would allow the applications of any StyleGAN-based editing approache~\cite{patashnik2021styleclip,shen2020interpreting} 
%are applicable for editing 
on the InD component such as changing facial expression, which is often desirable for user experiences. 
% Benefits
The advantages of our work are three-fold: 
% Better reconstruction
a) we achieve a higher-fidelity reconstruction by composition of InD and OOD components;
% Better editing
%b) by editing \emph{only} the InD part, we maintain the editability,
b) we retain the editability of pre-trained GANs by editing \emph{only} the InD content;
% the editability is well-preserved,
% Better 3D pose change
and 
c) by leveraging 3D-aware GANs, we can render the face from novel viewpoints.

We evaluate our method on challenging in-the-wild face images and videos, demonstrating improvement over previous state-of-the-art GAN inversion work on both reconstruction and editing quality.
%on face images and its potential for semantic video editing.
In addition, we demonstrate the usefulness of our method with 3D-aware editing applications, including %such as 
semantic editing, novel view synthesis, and OOD object removal.

We will release the code and data used in the paper.

% What are the specific contributions? Start with ``Our contributions are ..."
\topic{Our contributions.}
In summary, our contributions are:
\begin{itemize}[noitemsep,topsep=0pt, leftmargin=*]
\item We propose a 3D-aware GAN inversion approach to manipulate single images or monocular videos with out-of-distribution objects (e.g., accessories and heavy make-up). See results in Figure~\ref{fig:teaser}.
\item We incorporate composite volume rendering into 3D-aware GAN inversion.
\item Our method reconstructs 3D shapes of faces with OOD objects faithfully and demonstrates novel 3D-aware applications.  
\end{itemize}

%% file: figures/motivation.tex
\begin{figure}
\begin{center}
\centering

\frame{\includegraphics[trim=0 0 0 0, clip,width=.150\textwidth]{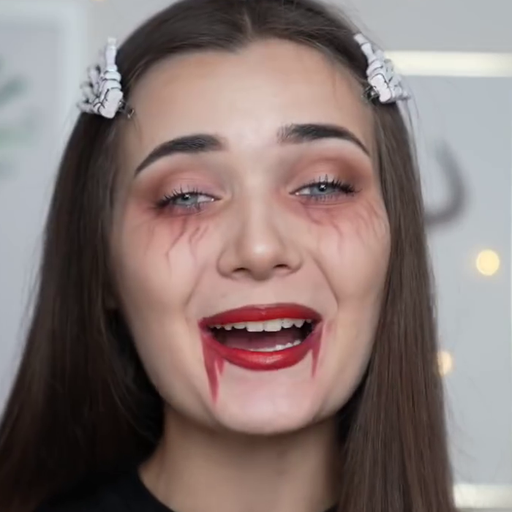}}\hspace{-20mm}\hfill
\mpage{0.01}{\raisebox{70pt}{\rotatebox{90}{\textit{More smile}}}}\hspace{1mm}
\frame{\includegraphics[trim=0 0 0 0, clip,width=.150\textwidth]{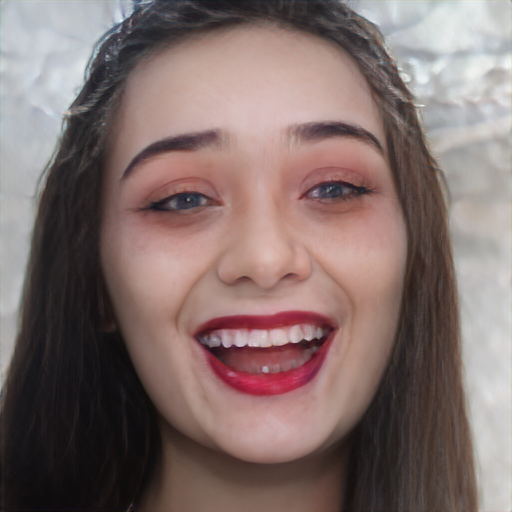}}\hspace{-20mm}\hfill
\frame{\includegraphics[trim=0 0 0 0, clip,width=.150\textwidth]{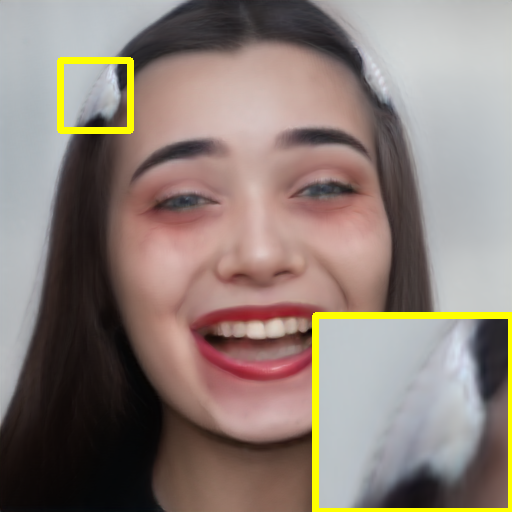}}\\

\vspace{-19mm}

\frame{\includegraphics[trim=0 0 0 0, clip,width=.150\textwidth]{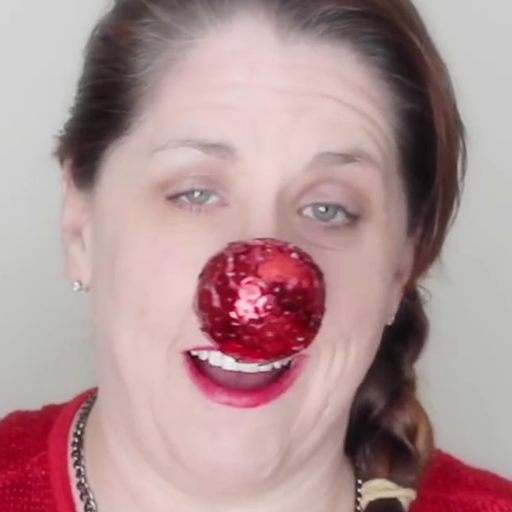}}\hspace{-20mm}\hfill
\mpage{0.01}{\raisebox{70pt}{\rotatebox{90}{\textit{Surprised}}}}\hspace{1mm}
\frame{\includegraphics[trim=0 0 0 0, clip,width=.150\textwidth]{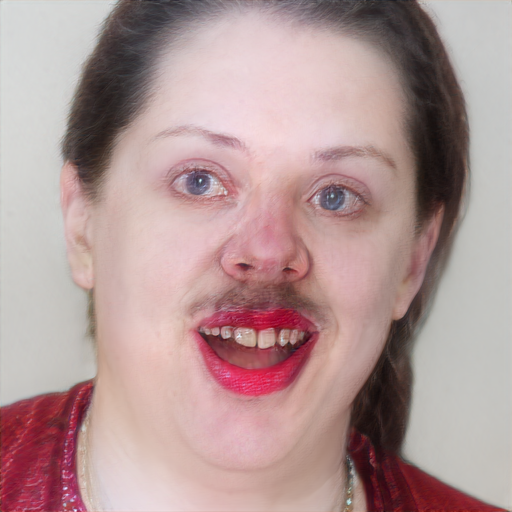}}\hspace{-20mm}\hfill
\frame{\includegraphics[trim=0 0 0 0, clip,width=.150\textwidth]{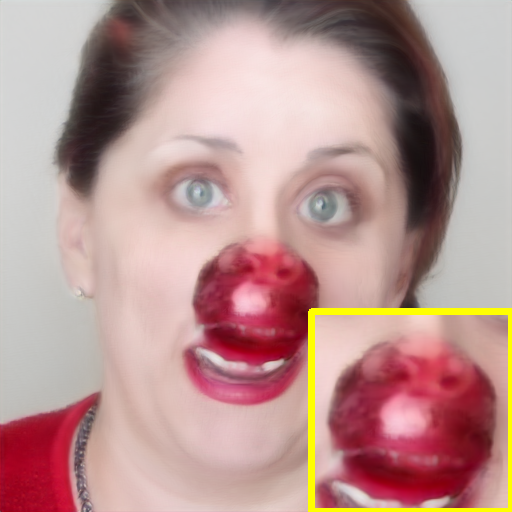}}\\

\vspace{-18mm}

\mpage{0.3}{\small{Input}}\hfill
\mpage{0.35}{\small{GOAE~\cite{Yuan_2023_GOAE}}} \hfill
\mpage{0.25}{\small{PTI~\cite{roich2022pivotal}}}

\caption{
\textbf{Limitations of the previous methods.} 
% Given an image with OOD objects, 
Existing GAN inversion techniques cannot deal with frames with OOD elements,
resulting in a poor \emph{reconstruction-editing balance}.
GOAE~\cite{Yuan_2023_GOAE} can produce faithful editing, but fails to preserve the identity of the input face.
PTI~\cite{roich2022pivotal} provides higher reconstruction fieldity, but the edibility suffers. 
}
\label{fig:motivation}
\end{center}
\vspace{-10mm}
\end{figure}

%% file: 2_related.tex
\section{Related Work}
\label{sec:related}
 
% Relationship examples:
% Similar:
% - Our work also adopt X ...
% - We address similar challenges.
% Different:
% - Our work differs in X ...
% - In contrast, we tackle ...

\topic{3D-aware GANs.}
StyleGANs~\cite{karras2019style,karras2020analyzing,karras2020training,karras2021alias} have achieved high-quality photorealistic 2D image generation and have been successfully applied to various image editing applications~\cite{shen2020interpreting,patashnik2021styleclip,gal2022stylegan,harkonen2020ganspace}.
Significant progresses have also been made to lift 2D image generation to 3D space, using various 3D representations, for both higher quality generation and to enable 3D-aware applications such as view synthesis~\cite{nguyen2019hologan,schwarz2020graf,chan2021pi,deng2022gram,gu2022stylenerf,orel2022styleSDF,chan2022efficient,sun2022ide,skorokhodov2022epigraf,gao2022get3d}.
These methods usually take a two-stage pipeline that renders a raw image (usually also with feature maps) in low resolution and then upsamples the rendered image to high resolution. 
% Relationship: our method is built upon EG3D
We leverage EG3D~\cite{chan2022efficient} as our generator architecture in this work.
% EG3D builds upon a StyleGAN2~\cite{karras2020analyzing} backbone architecture and nicely inherits the qualities of a well-behaved latent space that naturally allows effective GAN inversion and editing applications.

\topic{GAN inversion and editing.}
% What's going on in this topic?
% 2D GAN Inversions
GAN inversion has been widely studied for 2D GANs.
%They can be split into
These techniques can largely be categorized as 
% encoder-based
(a) encoder-based methods~\cite{luo2017learning,tewari2020stylerig,Nitzan2020FaceID,viazovetskyi2020stylegan2,richardson2021encoding,alaluf2021restyle,tov2021designing,chai2021latent,richardson2021encoding,tov2021designing} in which a neural network encoder is trained to project an input image to the latent space of the generator; 
% optimization-based
(b) optimization-based methods~\cite{raj2019gan,abdal2019image2stylegan,abdal2020image2stylegan++,huh2020ganprojection,tewari2020pie,gu2020image,collins2020editing,daras2020your} where the latent code is recovered via optimizing loss functions between the generator output and a target image; 
% hybrids
and (c) hybrid methods~\cite{zhu2020indomain,bau2020semantic,roich2022pivotal,alaluf2022hyperstyle} which combine both approaches. 
% 3D GAN
Some recent works have also investigated 3D-aware GAN inversion from a single image~\cite{xie2023hfgi3d,Yuan_2023_GOAE,lan2023e3dge,yin2023spi,trevithick2023real,lin20223dganinversion,sun2022ide} or a video~\cite{fruhstuck2023vive3d, zhang2022training}.
% Unlike previous methods focusing on single-image inversion, we consider multi-shot (or video) inversion. 
% With video input, our approach can better reconstruct 3D shapes.
As our experiments demonstrate, previous approaches have difficulty handling these challenging cases.
We propose a new mechanism to allow high-quality 3D-aware GAN inversion of \emph{out-of-distribution faces} even under significant occlusion.
% Editing
With our GAN inversion, we can modify the latent code to perform high-quality semantic image editing~\cite{shen2020interpreting,harkonen2020ganspace,patashnik2021styleclip,gal2022stylegan} or video editing~\cite{yao2021latent,tzaban2022stitch,xu2022temporally}.
% Relationship: our goal is also to do inversion and editing, but for out-of-distribution images.
% Our work uses an optimization-based approach for the GAN inversion, and we focus on the out-of-distribution videos.

\topic{GAN inversion for out-of-distribution (OOD) data.}
% What's going on in this topic?
% W+: better reconstruction
There have been attempts to invert out-of-distribution data to the GAN's latent space. 
Early work~\cite{abdal2019image2stylegan} proposes to project an image onto extended $\mathcal{W^{+}}$ space to achieve more accurate reconstruction.
% PTI: finetune G for better reconstruction
PTI~\cite{roich2022pivotal} finetunes generator with regularization for a lower distortion error. 
% StyleSpace: looking for feature maps
StyleSpace~\cite{wu2021stylespace} proposes to invert an image using StyleGAN's internal feature maps and tRGB blocks, which shows better reconstruction and disentanglement. 
% ChunkyGAN: 
Recently, ChunkyGAN~\cite{vsubrtova2022chunkygan} proposes to compose multiple generated images from multiple latent codes, with a set of segmentation masks to reconstruct an input image. 
% Relationship: same goal, but different 
With a similar goal in mind, we propose to leverage the radiance field of EG3D~\cite{chan2022efficient} and decompose the volumetric representation into an in-distribution part and an out-of-distribution part.
In contrast to ChunkyGAN~\cite{vsubrtova2022chunkygan} that models an image as a collection of \emph{2D} segments, we model the OOD and face directly in volumetric 3D representation and merge them with composite rendering.
% We propose reconstructing these two parts separately and composing them using a composite rendering scheme.

% \topic{3D representations.}
% Representing a scene or an object in 3D has been a classical problem. 
% % Explicit
% Explicit representations, like voxels~\cite{lombardi2019neural,sitzmann2019deepvoxels}, are fast but also suffer from memory overflow. 
% % Implicit
% Implicit representations~\cite{park2019deepsdf,sitzmann2020implicit,mescheder2019occupancy,mildenhall2020nerf}, on the other hand, shows better memory efficiency and achieves remarkable results recently, but they usually suffer from a slow rendering speed. 
% % Hybrid
% Meanwhile, combining both together, hybrid representations ~\cite{liu2020neural,peng2020convolutional,martel2021acorn,Chen2022ECCV,chan2022efficient} gain a balance between memory efficiency and rendering speed. 
% % Relationship
% In our work, we use tri-planes~\cite{Chen2022ECCV,chan2022efficient}, a hybrid representation, as our 3D representation.

\topic{Composite neural radiance fields.}
% What's going on in this topic?
% NeRF
Neural Radiance Fields (NeRFs)~\cite{mildenhall2020nerf} have shown impressive view synthesis results. 
% However, the vanilla NeRF is only used for a single, static scene.
% Composable NeRFs
Recently, it has been shown that 3D scenes can be decomposed into different NeRFs.
When multiple radiance fields are built, one can compose them together using a composite rendering manner~\cite{martin2021nerf,gao2021dynamic,yang2021learning,wu2022d}.
% Relationship
EG3D~\cite{chan2022efficient} uses the tri-plane representation to generate 3D objects from the latent code.  
We adopt the idea of composite volume rendering to address the out-of-distribution 3D GAN inversion problem.
Specifically, we split the in-distribution and out-of-distribution parts in the tri-plane 3D representation and compose them during volume rendering.

%% file: 3_method.tex
% Images
\def\D{\altmathcal{D}}
\def\I{\altmathcal{I}}
\def\O{\altmathcal{O}}
\def\res{\altmathcal{R}}

% Vectors
\def\b{\mathbfit{b}}
\def\c{\mathbfit{c}}
\def\d{\mathbfit{d}}
\def\o{\mathbfit{o}}
\def\p{\mathbfit{p}}
\def\t{\mathbfit{t}}
\def\x{\mathbfit{x}}
\def\z{\mathbfit{z}}

% Matrices
\def\K{\mathbfit{K}}
\def\R{\mathbfit{R}}

% Functions
\def\ang{\phi}
\def\dehom{\mu}
\def\proj{\pi}
\def\sigmoid{S}
\def\vis{\nu}
\def\r{\mathbfit{r}}

% Bracketed
\def\bp{(\p\!)} % (p)
\def\bt{(t\!)} % (p)
\def\bx{(\x\neg)} % (x)

% Subsets
\def\ok{\o_{\neg k}}
\def\tk{\t_{\neg k}}
\def\wk{w_{\neg k}}
\def\xi{\x_{\neg i}}
\def\zk{\z_{\neg k}}
\def\Kk{\K_{\neg k}}
\def\Rk{\R_{\neg k}}

% Helpers
\def\ng{\hspace{-0.1mm}}
\def\neg{\hspace{-0.2mm}}
\def\pos{\hspace{0.2mm}}

% Create \overrightharpoon, as a replacement for \overrightvector.
\makeatletter
\newcommand*\MY@rightharpoonupfill@{%
    \arrowfill@\relbar\relbar\rightharpoonup
}
\newcommand*\overrightharpoon{%
    \mathpalette{\overarrow@\MY@rightharpoonupfill@}%
}
\makeatother

% A scalable sum symbol. Use like this: \nsum[0.8]
\newlength{\depthofsumsign}
\setlength{\depthofsumsign}{\depthof{$\sum$}}
\newcommand{\nsum}[1][1.4]{% only for \displaystyle
    \mathop{%
        \raisebox
            {-#1\depthofsumsign+1\depthofsumsign}
            {\scalebox
                {#1}
                {$\displaystyle\sum$}%
            }
    }
}

%%%%%%%%%%
%%%%%%%%%%

\section{3D-aware GAN: EG3D}
\label{sec:eg3d_arch}

% \subsection{Neural Radiance Fields (NeRFs)}\label{sec:nerf_intro}
% % Brief intro to NeRF
% % Why we need to build a NeRF?
% Our goal is to leverage the composability of Neural Radiance Fields~\cite{martin2021nerf} (NeRFs) to reconstruct the OOD object and in-distribution (InD) face, respectively. 
% % What is a NeRF?
% A NeRF is an implicit 3D representation with a differentiable and continuous function $D$ (either an MLP or voxel grid features), that takes a 3D position $\mathbf{x} \in \mathbb{R}^3$ and a view direction $\mathbf{d} \in \mathbb{R}^3$ as the input,
% and outputs pointwise RGB color $\mathbf{c} \in \mathbb{R}^3$ and density $\sigma$.
% \begin{equation}
%     \mathbf{c}(\mathbf{x}, \mathbf{d}), \sigma(\mathbf{x}) = D(\mathbf{x}, \mathbf{d}) \,.
% \end{equation}

% % How it works?
% To compute the color of a pixel, a ray $\mathbf{r}(t_k)=\mathbf{o}+t_k\mathbf{d}$ is cast through the camera origin in the direction $\mathbf{d}$. 
% We then compute the predicted color by volume rendering~\cite{max1995optical}:
% \begin{equation}
%     \mathbf{C}(\mathbf{r}) = \sum^K_{k=1}T(t_k) \alpha(\sigma(t_k)\delta_k) \mathbf{c}(t_k) \,,
% \end{equation}
% where $T(t_k) = \exp(-\sum^{k-1}_{k^{\prime}=1} \sigma(t_{k^{\prime}}) \delta_{k^{\prime}})$,
% $\alpha=1-\exp(-x)$, and $\delta_k = t_{k+1}-t_k$ is the distance between two 3D points. 

% \subsection{3D-aware GAN: EG3D}\label{sec:eg3d_arch}
% 3D representation + SR module
% What is EG3D?
We choose EG3D~\cite{chan2022efficient}, which consists of a tri-plane representation and a super-resolution (SR) module, as our 3D-aware GAN.

% How it works?
% Low-res from tri-plane + rendering
\topic{Neural rendering at low resolution.} 
Given a latent code $z \in \mathbb{R}^{512}$ (or $w \in \mathbb{R}^{14\times512}$) and camera parameters $p$, EG3D first generates a corresponding tri-plane $\mathbf{T} \in \mathbb{R}^{256\times256\times32\times3}$.
For each pixel, a ray $\mathbf{r}$ is cast, and points are sampled along the ray. 
Unlike the positional encoding~\cite{martin2021nerf,tancik2020fourier} for each point in NeRFs~\cite{martin2021nerf}, EG3D projects each point onto tri-plane $\mathbf{T}$ and retrieves features from three planes via bilinear interpolation.
These features are then aggregated by summation, and fed into the decoder $D$ (\ie an MLP) to predict the color and density.
Volume rendering~\cite{max1995optical} is then performed to compute the final color for each pixel. 
To this end, a raw RGB image with a 32-channel feature in a low resolution (\eg $128\times128$) is generated.

% High-res from SR
\topic{Super-Resolution (SR).} To gain high-resolution outputs, EG3D later uses an SR module that inputs the raw image and the 32-channel feature as the input and yields a high-resolution RGB image (\eg $512\times512$).
% Why we need it?
We build our approach upon EG3D due to its rendering efficiency compared to other alternatives~\cite{orel2022styleSDF,gu2022stylenerf}.
% Compared to other 3D representations~\cite{orel2022styleSDF,gu2022stylenerf} in 3D GANs, EG3D is more efficient during the rendering as it uses a hybrid representation (\ie tri-plane). 

\section{Method}
\label{sec:method}
\input{figures/method_overview.tex}
% Brief I/O statement
% Given an aligned, monocular face video $\mathbf{V}=[\mathbf{I}_1,\cdots, \mathbf{I}_t, \cdots, \mathbf{I}_N]$ with $N$ frames, 
% and binary masks $[\mathbf{M}_1, \cdots, \mathbf{M}_t, \cdots, \mathbf{M}_N]$ of the out-of-distribution (OOD) content for each frame, 
%our goal is to reconstruct the face video and edit it. 
Given an aligned face input image $\mathbf{I}$, or a monocular face video $\mathbf{V}=[\mathbf{I}_1,\cdots, \mathbf{I}_t, \cdots, \mathbf{I}_N]$ with $N$ frames, we aim to reconstruct the input with EG3D inversion and perform face editing. 
For simplicity, we use $\mathbf{I}_t$ to represent a frame, either from a single input image or a sampled frame from a video. 
If only one frame exists, then $N=1$.

% Approach idea
We present the high-level overview in Figure~\ref{fig:method_overview}.
% As shown in Figure~\ref{fig:method_overview}, 
% Each time we sample one single frame $\mathbf{I}_t$ from the video $\mathbf{V}$. Our method is unsupervised, \ie, no need for segmentation masks like previous methods~\cite{gao2021dynamic,yang2021learning}.
% Split
We build \emph{two} neural radiance fields (NeRFs)~\cite{mildenhall2020nerf}, 
% In-distribution
one for \emph{in-distribution} (InD) face (Section~\ref{sec:id_inv}), 
% OOD
and the other one for \emph{out-of-distribution} (OOD) object (Section~\ref{sec:ood_inv}), using tri-plane representations~\cite{chan2022efficient}. 
The OOD object, for example, can be a non-face object with a rigid shape or heavy makeup with a complicated texture. 
% We leverage EG3D's architecture since it consists of a neural rendering module with a tri-plane representation and a super-resolution module. 
% Compose
Next, we combine two radiance fields (Section~\ref{sec:composite_render}) to reconstruct the low-resolution frame.
% SR
Finally, we finetune the super-resolution module of EG3D to get the high-resolution output (Section~\ref{sec:sr}).
% Editing
After training the radiance fields, we can edit the face image or video (Section~\ref{sec:edit}). 

\subsection{In-distribution GAN inversion}\label{sec:id_inv}
% As mentioned in Section~\ref{sec:eg3d_arch}, 
% EG3D~\cite{chan2022efficient} $G$ first takes a latent code $w \in \mathbb{R}^{14\times512}$ as the input to generate a tri-plane representation $\mathbf{T}^I \in \mathbb{R}^{256\times256\times32\times3}$.
% Then the features are sampled from $\mathbf{T}^I$ and fed into an MLP decoder $D^I$.
% The decoder predicts the colors $\mathbf{c}^{I}$ and densities $\sigma^I$ for the 3D coordinates.
% Next, volume rendering~\cite{max1995optical} is used to generate a low-resolution output ($128\times128$).

\topic{Formulation.}
% Basic idea
Since a pretrained EG3D already has prior knowledge of faces, we directly leverage its latent space and perform a regular 3D GAN inversion~\cite{lin20223dganinversion,xia2022gan} for the in-distribution part.
% The in-distribution inversion is similar to the normal 3D GAN inversion~\cite{lin20223dganinversion,xia2022gan}.
% except we are inverting a face \emph{video} (or multiple images of the same identity).
For a single frame case, we optimize a latent code $w_t$ such that it can reconstruct the input frame $\mathbf{I}_t$.
For a video, we invert all the frames at the same time. Please refer to the supplementary material for more details. 
For camera parameters $p_t \in \mathbb{R}^{25}$, we obtain them by using an off-the-shelf pose detector~\cite{deng2019accurate}, following~\cite{chan2022efficient,lin20223dganinversion}.
\topic{Optimization.} 
To represent the InD faces with $\mathbf{T}^I$, our insight is to keep the latent code $w_t$ in distribution as much as possible. 
To this end, we use a regularization term to keep $w_t$ within its pre-trained distribution through GAN training. 
\begin{equation} \label{eq:latent_reg}
    % \scriptsize
    \mathcal{L}_{w} (w_t)  = ||w_t - \bar{w}||^2_2 \,,
\end{equation}
where $\bar{w}$ is the mean latent code computed over 10,000 sampled latent codes. 

We also use a another regularization term adopted from~\cite{tov2021designing} to constrain the variation among style vectors in $w$: $\mathcal{L}_{\Delta} (w_t) = \sum^{13}_{i=1}||\Delta_i||^2_2$,
% \begin{equation} \label{eq:latent_delta}
%     % \scriptsize
%     \mathcal{L}_{\Delta} (w_t) = \sum^{13}_{i=1}||\Delta_i||^2_2 \,,
% \end{equation}
given a latent code $w = (w_0, w_0 + \Delta_1, ..., w_0 + \Delta_{13}) \in \mathbb{R}^{14\times512}$.  
This regularization term preserves the editability of the optimized latent code~\cite{tov2021designing}. 

% \topic{Optimization.} Our optimization objective for this stage is 
% \begin{equation}\label{eqn:goal_id}
%     % \scriptsize
%         \begin{split}
%         % w^{*}_t, p^{*}_t & = \argmin_{w_t, p_t} \mathcal{L}^{I}_t  = 
%         w^{*}_t &= \argmin_{w_t} \mathcal{L}^{I}_t  = 
%         \argmin_{w_t} \frac{1}{||\mathbf{m}_t||_1}||\mathbf{m}_t \odot (\mathbf{x} - \mathbf{\hat{x}})||^2_2 \\ 
%         &+ \mathcal{L}_{\mathrm{mLPIPS}}(\mathbf{x}, \mathbf{\hat{x}}, \mathbf{m}_t)
%         + \lambda_{\Delta} \mathcal{L}_{\Delta} (w_t)  \,,
%     \end{split}
% \end{equation}
% where $w_t$ is the latent code at time $t$, $\mathbf{x}$ is the input frame $\mathbf{I}_t$, $\mathbf{\hat{x}}$ is generation output $G(w_t, p_t)$, $\mathbf{m_t}=1-\mathbf{M}_t$, and $\mathcal{L}_{\mathrm{mLPIPS}}(x, \hat{x}, \mathbf{m})$ is the masked version of LPIPS loss~\cite{zhang2018perceptual}. 
% Here, $p_t$ is the camera parameters estimated by~\cite{deng2019accurate}.
% The masked LPIPS only considers features inside the mask $\mathbf{m}$.  
% $\mathcal{L}_{\Delta}(w)= \sum^{13}_{i=1}||\Delta_i||^2_2$ is a regularization loss adopted from~\cite{tov2021designing}, used to constrain the variation among style vectors in $w$

% given a latent code $w = (w_0, w_0 + \Delta_1, ..., w_0 + \Delta_{13}) \in \mathbb{R}^{14\times512}$. 
% By minimizing the variation, we push $w \in \mathcal{W}^{+}$ to be closer to $\mathcal{W}$ space, which has better editability.

\subsection{Modeling out-of-distribution contents}\label{sec:ood_inv}
For an OOD object, a pre-trained EG3D usually cannot model it well with its prior distribution.
%EG3D has no prior knowledge about it. 
We therefore use an additional tri-plane $\mathbf{T}^O$ to represent the out-of-distribution content. 
% Challenge
%However, , especially for a video, 
One additional challenge is that, while dealing with video, the OOD object may not be static across different frames, therefore could not be well reconstructed with a static radiance field. 
% How to deal with it
Therefore, in addition to $\mathbf{T}^O$, we use a per-frame latent code $\mathbf{\phi}_t \in \mathbb{R}^{32}$ for each frame to represent the out-of-distribution object across the temporal domain. 
% Additional details
Both $\mathbf{T}^O$ and $\mathbf{\phi}_t$ are randomly initialized from a normal distribution. 

\topic{Formulation.} 
% Prediction.
The out-of-distribution decoder $D^O$ takes a tuple $(\mathbf{T}^O(t_k), \mathbf{\phi}_t) \in \mathbb{R}^{64}$ as the input, and outputs color $\mathbf{c}^O \in \mathbb{R}^3$, density $\sigma^O \in \mathbb{R}$, and blending weight $b \in [0, 1]$. 
\begin{equation} \label{eqn:ood_decoder}
    (\mathbf{c}^O, \sigma^O, b) = D^O(\mathbf{T}^O(t_k), \mathbf{\phi}_t; \theta_{D^O}) \,,
\end{equation}
where $\mathbf{T}^O(t_k) \in \mathbb{R}^{32}$ is the aggregated features obtained by projecting 3D coordinate $t_k$ onto each of the three feature planes via bilinear interpolation, then aggregated via summation~\cite{chan2022efficient}.
The decoder $D^O$ is an MLP with weights of $\theta_{D^O}$.
% Rendering.
To compute the color of a pixel at time $t$, we use the volume rendering integral along the ray $\mathbf{r}$:
\vspace{-2mm}
\begin{equation}
    \mathbf{C}^O(\mathbf{r}) = \sum^{K}_{k=1} T(t_k) \alpha^O(\sigma^O(t_k) \delta_k) \mathbf{c}^O(t_k) \,,
\end{equation}
where $T(t_k) = \exp(-\sum^{k-1}_{k^{\prime}=1} \sigma(t_{k^{\prime}}) \delta_{k^{\prime}})$,
$\alpha=1-\exp(-x)$, and $\delta_k = t_{k+1}-t_k$ is the distance between two 3D points. 
\vspace{-4mm}
% where
% \begin{equation}
%     T(t_k) = \exp(-\sum^{k-1}_{k^{\prime}=1} \sigma(t_{k^{\prime}}) \delta_{k^{\prime}} ) \,,
% \end{equation}
% $\delta_k = t_{k+1} - t_{k}$ is the distance between two points, $\alpha(x)=1-\exp(-x)$.

% Training.
% \topic{Optimization.} The optimization objective is:
% \begin{equation}\label{eqn:goal_ood}
%     \mathbf{T}^{O*}, \theta^{*}_{D^O}, \mathbf{\phi}^{*}_t = \argmin_{\mathbf{T}^O, \theta_{D^O}, \mathbf{\phi}_t} \mathcal{L}^{O}_t \,,
% \end{equation} 
% where
% \begin{equation}\label{eqn:loss_ood}
%     \mathcal{L}^{O}_t = \sum_{ij} ||(\mathbf{C}^O_t(\mathbf{r}_{ij}) - \mathbf{C}^{GT}(\mathbf{r}_{ij})) \cdot \mathbf{M}_t(\mathbf{r}_{ij})||^2_2 \,,
% \end{equation}
% \begin{equation}\label{eqn:goal_ood}
%     \scriptsize
%     \begin{split}
%         & \mathbf{T}^{O*}, \theta^{*}_{D^O}, \mathbf{\phi}^{*}_t  = \argmin_{\mathbf{T}^O, \theta_{D^O}, \mathbf{\phi}_t} \mathcal{L}^{O}_t \\
%         & = \argmin_{\mathbf{T}^O, \theta_{D^O}, \mathbf{\phi}_t} \sum_{ij} ||(\mathbf{C}^O_t(\mathbf{r}_{ij}) - \mathbf{C}^{GT}(\mathbf{r}_{ij})) \cdot \mathbf{M}_t(\mathbf{r}_{ij})||^2_2   \,,
%     \end{split}
% \end{equation} 
% $\mathbf{C}^{GT}(\mathbf{r}_{ij})$ is the ground-truth color at pixel $(i,j)$.
% \textcolor{red}{$\mathbf{C}^{GT}(\mathbf{r}_{ij}))$ is the ground-truth color at pixel $(i,j)$.} 

\subsection{Composite volume rendering}
\label{sec:composite_render}
% Overview
Now, with both InD and OOD radiance fields, we can combine them using the blending weight $b$ from Eqn.~\ref{eqn:ood_decoder}.

\topic{Formulation.} 
We compose two radiance fields together by
\vspace{-5mm}
\begin{equation} \label{eq:composite_rendering} 
    % \scriptsize	
    \begin{split}
        \mathbf{C}^{C}(\mathbf{r}) &= \sum^{K}_{k=1} T^{C}(t_k) \Bigl( b \alpha^O(\sigma^O(t_k) \delta_k) \mathbf{c}^O(t_k)  \\
        &+ (1 - b) \alpha^I(\sigma^I(t_k) \delta_k) \mathbf{c}^I(t_k) \Bigr) \,,
    \end{split}
\end{equation}
where $T^{C}(t_k) = \exp(-\sum^{k-1}_{k^{\prime}=1} (\sigma^O + \sigma^I) \delta_{k^{\prime}} ) \,.$

\topic{Optimization.} 
The goal is 
\begin{equation}\label{eqn:goal_comp}
    \scriptsize
    \begin{split}
         w^{*}_t, \mathbf{T}^{O*}, \theta^{*}_{D^O}, \mathbf{\phi}^{*}_t & = \argmin_{w_t, \mathbf{T}^O, \theta_{D^O}, \mathbf{\phi}_t} \mathcal{L}^C_t   \\
        &= \argmin_{w_t, \mathbf{T}^O, \theta_{D^O}, \mathbf{\phi}_t} \sum_{ij}||C^C(\mathbf{r}_{ij}) - C^{GT}(\mathbf{r}_{ij})||^2_2   \\
        &+ \lambda_{b}\mathcal{L}_b(\mathbf{r}_{ij}) 
        % + \sum_{\mathbf{M}_t(ij) \neq 1}\lambda_{spar}\mathcal{L}_{spar} 
        + \mathcal{L}_{LPIPS}(\mathbf{I}^C_{LR}, \mathbf{I}_{LR})
        \,,
    \end{split}
\end{equation}
where $\mathcal{L}_{LPIPS}$ is the LPIPS loss~\cite{zhang2018perceptual}, $\mathbf{I}^C_{LR}$ is the composite rendered image at low resolution ($128\times128$), $\mathbf{I}_{LR}$ is the ground truth image also at $128\times128$.
% $\mathcal{L}_{spar}$ is a sparsity loss used to suppress the blending weights for the OOD pixels outside the mask. 
% We use this regularization in case the OOD radiance field dominates, which makes the editing trivial, 
% since we rely on the in-distribution part for the editing. 
% \begin{equation}\label{eqn:spar_loss}
%     \scriptsize
%     \mathcal{L}_{spar}(\mathbf{r}) = \sum^K_{k=1} \mathcal{L}_1(b(t_k)) \,, 
% \end{equation}
The weight regularizer $\mathcal{L}_b$ is adopted from~\cite{wu2022d}, used to penalize the blending weight $b$ if it is not closer to 0 or 1:
\begin{equation}\label{eqn:blendw_loss}
    \scriptsize
    \mathcal{L}_b(\mathbf{r}) = \sum^K_{k=1} H_b(b(t_k)) \,,
\end{equation}
where $H_b(x) = -(x\log(x) + (1 - x)\log(x))$ is binary entropy.
The reason behind Eqn.~\ref{eqn:blendw_loss} is that objects cannot co-occupy \emph{the same spatial location}.
The entropy loss facilitates a cleaner decomposition: encouraging an object to be either in-distribution (\ie $b \rightarrow 0$) or out-of-distribution (\ie $b \rightarrow 1$).

However, it is ill-posed to build its 3D geometry accurately, given only a single-frame input, even with a pre-trained 3D-aware generator.
Therefore, for a single image only, we also introduce a depth regularization term:
\begin{equation}
    \mathcal{L}_{\mathcal{D}} = ||\mathcal{D}^{C} - \mathcal{D}^{Reg} ||_1  \,
\end{equation}
where $\mathcal{D}^{C}$ is the rendering depth map from composite rendering, and $\mathcal{D}^{Reg}$ is a rescaled depth map from MiDaS~\cite{Ranftl2022midas}.

% In practice, we first optimize for Eqn.~\ref{eqn:goal_id} only. 
% We then jointly optimize for Eqn.~\ref{eqn:goal_ood} and Eqn.~\ref{eqn:goal_comp}. 
% It takes ~3.5 hours to obtain an accurate reconstruction for a video of 200 frames, on an NVIDIA RTX A4000 GPU.
% \subsection{Total loss functions}
\subsection{Low-resolution reconstruction}
\label{sec:low_res_rec}
In practice, we jointly optimize for ${w_t}$ (yielding $\mathbf{T}^{I}$), $\mathbf{T}^{O}$, $\theta_{D^O}$, $\mathbf{\phi}_t$, following Section~\ref{sec:id_inv} to Section~\ref{sec:composite_render}. 
Our total loss function is
\begin{equation} \label{eq:low_res_loss}
    \mathcal{L}^{LR} = \sum_{t=1}^{N} \mathcal{L}^{C}_{t} + \lambda_{\Delta}\mathcal{L}_{\Delta} + \lambda_{w} \mathcal{L}_{w} + (\lambda_{\mathcal{D}} \mathcal{L}_{\mathcal{D}}) \,,
\end{equation}
where $\mathcal{L}^{C}_{t}$ is from Eqn.~\ref{eq:composite_rendering}, latent variation regularizer $\mathcal{L}_{\Delta}$ from~\cite{tov2021designing}, and $\mathcal{L}_{w}$ from Eqn.~\ref{eq:latent_reg}, respectively, 
$\lambda_{\Delta}$ is the weight for $\mathcal{L}_{\Delta}$, $\lambda_{w}$ is the weight for $\mathcal{L}_{w}$, and $\lambda_{\mathcal{D}}$ is the weight for $\mathcal{L}_{\mathcal{D}}$. 
We only use $\mathcal{L}_{\mathcal{D}}$ for single image input.

\subsection{Super-Resolution}\label{sec:sr}
% why
After training in Section~\ref{sec:id_inv},~\ref{sec:ood_inv},~\ref{sec:composite_render} and~\ref{sec:low_res_rec}, 
we can get reconstruction $\mathbf{I}^C_{LR}$ in low resolution ($128\times128$).
We observe that using the pretrained super-resolution (SR) module cannot generate a satisfying high-resolution output, as shown in Figure~\ref{fig:sr}, due to the new OOD tri-plane $\mathbf{T}^{O}$.
Therefore, we finetune only the SR module in $G$ for higher resolution at $512\times512$.
\input{figures/sr_comp.tex}

\topic{Optimization.} The loss function is that
\begin{equation}\label{eqn:loss_sr}
    \mathcal{L}^{SR}(\mathbf{x}, \mathbf{\hat{x}}) = ||\mathbf{x} - \mathbf{\hat{x}}||^2_2 + \mathcal{L}_{LPIPS}(\mathbf{x}, \mathbf{\hat{x}}) \,,
\end{equation}
where $\mathbf{x} = \mathbf{I}_t$ and $\mathbf{\hat{x}} = SR(\mathbf{I}^C_{LR})$.

\subsection{Editing}\label{sec:edit}
After the reconstruction, we can modify the latent code $w_t$ to perform various semantic editing tasks. 
With explicit decomposition, the OOD contents do not interfere with the semantic editing capability of in-distribution components.
% Because we split the image into two radiance fields, out-of-distribution has little to do with semantic editing,
% during the editing, we only edit the latent code to get edited images. 
Here, any existing GAN-based editing approaches can be used.
We use InterfaceGAN~\cite{shen2020interpreting} and StyleCLIP~\cite{patashnik2021styleclip}.
% \input{figures/editing_overview.tex}
% Figure~\ref{fig:editing}.

% \subsection{Post-processing}\label{sec:postprocess}
% We can put the aligned frames back into the original input video as an optional step. 
% We use a face parsing algorithm~\cite{yu2021bisenet} to get the face segmentation 
% and paste the face back to the input frame with a Gaussian Blur to smooth the boundary.
% Finally, we paste the cropped, aligned frame into the full-resolution input video.

%% file: figures/method_overview.tex
\begin{figure*}
\begin{center}
\centering
\includegraphics[trim=0 0 0 0, clip,width=0.9\textwidth]{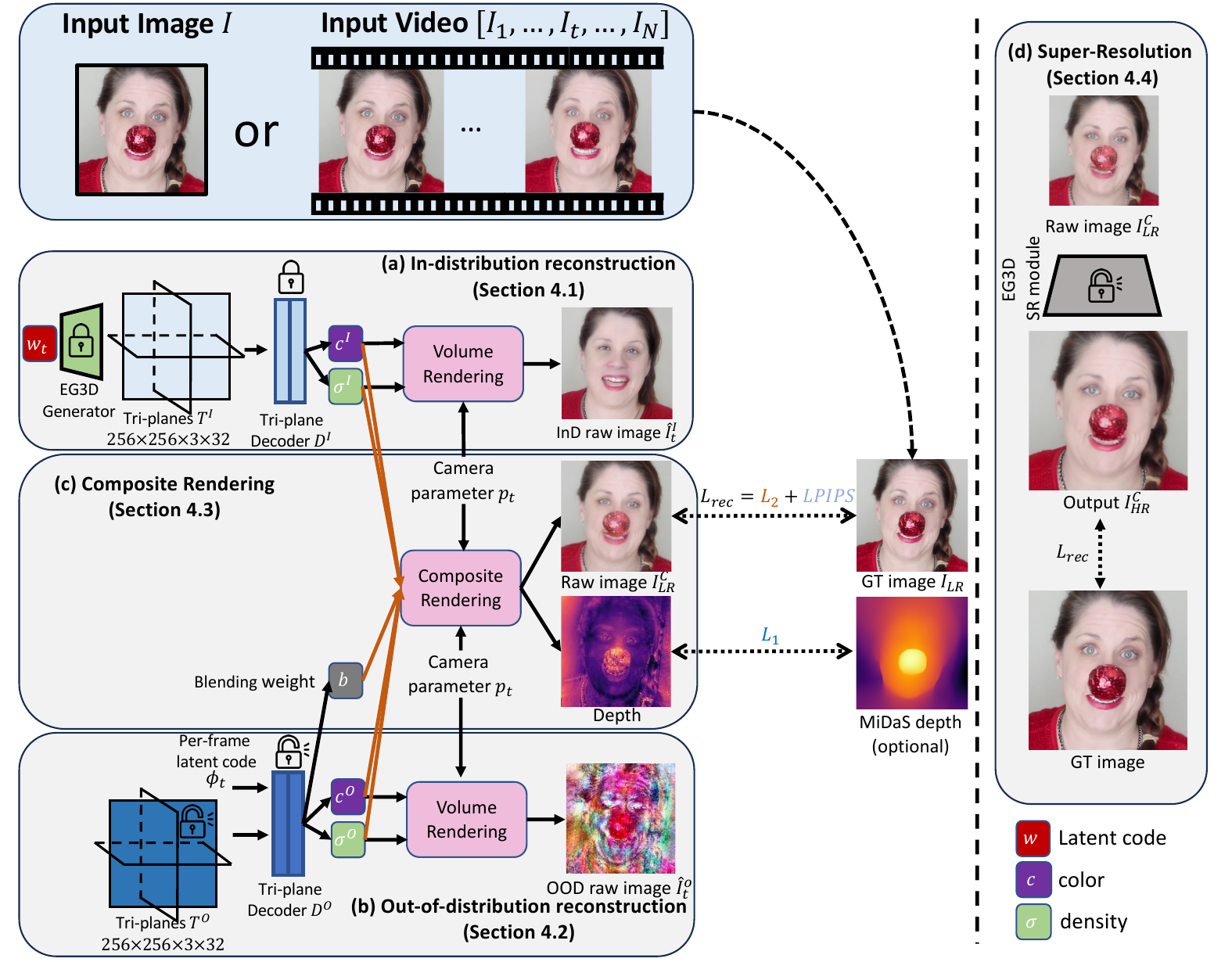}
\vspace{-3mm}
\caption{
\textbf{Overview of our method.} 
Given a potrait image or a monocular portrait video, we use two radiance fields to represent 
(a) \textit{in-distribution (InD)}  face, and (b) \textit{out-of-distribution (OOD)} item. 
(a) \textbf{InD reconstruction} is the \textit{GAN inversion} for the in-distribution natural face. 
We apply GAN inversion by using pre-trained EG3D model $G$ to the frame, where the pre-trained tri-plane generator and tri-plane decoder $D^I$ are kept frozen. 
(b) For \textbf{OOD} item, we propose to model them with a separate radiance field represented by an additional tri-plane $\mathbf{T}^O$. 
During the training process, we optimize the tri-plane $\mathbf{T}^O$, a per-frame latent code $\mathbf{\phi}_t$, and a new decoder $D^O$.
The decoder takes as input tri-plane features $\mathbf{T}^O$ and $\mathbf{\phi}_t$ and outputs color $\mathbf{c}^O$, density $\sigma^O$, and blending weight $b$. 
(c) \textbf{Composite Rendering} compose the \textit{InD} and \textit{OOD} radiance fields together by using a composite rendering scheme (Section~\ref{sec:composite_render}). 
(d) Finally, we finetune the \textbf{Super-Resolution} module in $G$ to achieve a better output in the high resolution.
After training, we can perform various semantic edits and free-view rendering, while preserving the face identity and the OOD components. 
}
\label{fig:method_overview}
\end{center}
\vspace{-8mm}
\end{figure*}

%% file: figures/sr_comp.tex
\begin{figure}
\begin{center}
\centering

\frame{\includegraphics[trim=0 0 0 0, clip,width=.15\textwidth]{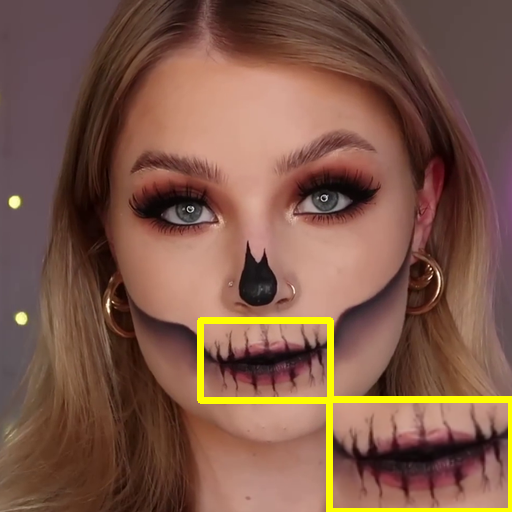}}\hfill
\frame{\includegraphics[trim=0 0 0 0, clip,width=.15\textwidth]{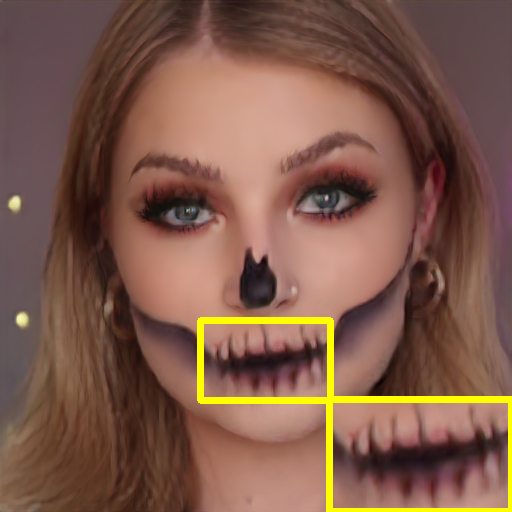}}\hfill
\frame{\includegraphics[trim=0 0 0 0, clip,width=.15\textwidth]{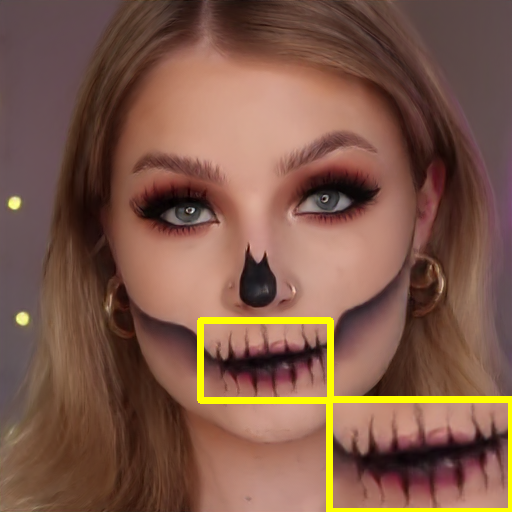}}\\

\mpage{0.3}{\small{Recon. Target}}\hfill
\mpage{0.3}{\small{w/o finetuning SR module}}\hfill
\mpage{0.3}{\small{w/ finetuning SR module}} \\

\caption{
\textbf{The effect of finetuning SR module.} 
Without finetuning the SR module, the high-resolution output is blurry. 
}
\label{fig:sr}
\end{center}
\vspace{-6mm}
\end{figure}

%% file: 4_result.tex
\section{Experimental Results}
\label{sec:result}

\subsection{Experimental Setup}
\label{sec:setup}
\input{figures/qualitative_comp_rec.tex}

\topic{Dataset.} 
To evaluate how our approach works on data with out-of-distribution components, we collected a dataset of 20 online videos with challenging and diverse appearances. The OOD content contains heavy make-up and occluding objects (\eg facial masks and large glasses).
For the single-image inversion method, we use the first frame of each video.
For video inversion, we use all the frames.
For the face alignment, we use 3DDFA-v2~\cite{guo2020towards} to obtain the 68-point landmarks and smooth them across the frames using a sliding window for stabler cropping. 
After that, we convert the landmarks to EG3D's 5-point landmarks and crop the face out of the input frame. 
% For the segmentation masks, we manually label the first frame and then use an off-the-shelf tracking algorithm~\cite{cheng2022xmem} to get the masks for the rest of frames. 
% For the face alignment, we use Deep3DFaceReconstruction~\cite{deng2019accurate} to get the landmarks, and smooth them across the frames using a sliding window for a stabler cropping. 

\topic{Hyperparameters.} 
We use the Adam optimizer~\cite{kingma2014adam} for all our experiments. 
For in-distribution inversion (Secion~\ref{sec:id_inv}), we optimize for 200 epochs with a learning rate of $1\times10^{-3}$, $\lambda_{\Delta}$ = $1\times10^{-3}$.
For the out-of-distribution and composite rendering (Secion~\ref{sec:ood_inv},~\ref{sec:composite_render}), we run the optimization for 10,000 iterations with a learning rate of $5\times10^{-3}$, $\lambda_b=1$, $\lambda_{w}=1$, and $\lambda_{\mathcal{D}}=0.1$ if applicable. 
For the SR module (Section~\ref{sec:sr}), we finetune the module for 100 epochs with a learning rate of $1\times10^{-3}$.

\topic{Metrics.} 
We evaluate our approach from 
1) reconstruction accuracy and 2) editability to validate the reconstruction-editability trade-off. 
For the reconstruction accuracy, we report LPIPS~\cite{zhang2018perceptual}, PSNR, SSIM and ID similarity~\cite{deng2019arcface}.
For editability, we follow~\cite{roich2022pivotal,vsubrtova2022chunkygan} and evaluate identity preservation after applying the editing direction. 
More specifically, we use ArcFace~\cite{deng2019arcface} to compute the similarity between the inverted and edited results. 

\topic{Baselines for evaluation.}
We compare our method extensively with several previous arts.
For optimization-based methods, we compare with HFGI3D~\cite{xie2023hfgi3d}, PTI~\cite{roich2022pivotal}, $\mathcal{W}{+}$, and $\mathcal{W}$ optimization.
For videos only, we also include VIVE3D~\cite{fruhstuck2023vive3d}.
We compare the encoder-based method with GOAE~\cite{Yuan_2023_GOAE} and IDE-3D~\cite{sun2022ide} encoder.
We treat $\mathcal{W}{+}$ optimization as an ablated version of our method \emph{without OOD triplane}.
The recent work in~\cite{trevithick2023real} showcases encoder-based 3D GAN inversion, focusing on real-time inference.
% We are also aware of another work~\cite{trevithick2023real} that can deal with some OOD images, focusing on real-time inference.
However, their method relies on a frozen EG3D and does not explicitly model the OOD components. 
% Our method, instead, is an optimization-based approach without additional large-scale training. 
We do not compare with it as the code is not publicly available.
% We do not include the experimental comparison with it since the code is not released. 
\vspace{-1mm}
\subsection{Quantitative results}\label{sec:quanitative}
% \yiran{Talk a little bit high-level for each baseline. 
% Specify W+. 
% }
\topic{Reconstruction.} 
% Setup
We compare the reconstruction accuracy of our approach with all baselines and report the results in Table~\ref{tab:quant_inv}. 
For PTI, we first perform a $\mathcal{W{+}}$ inversion with a learning rate of $1 \times 10^{-3}$ and 200 epochs, and then finetune the generator for 200 epochs with a learning rate of $3\times10^{-5}$.
For $\mathcal{W{+}}$ and $\mathcal{W}$ optimization, we use a learning rate of $1\times10^{-3}$ and optimize for 200 epochs.
For GOAE and IDE-3D, we use their encoder directly for the inversion. 
\input{tables/method_comp_inv.tex}

\input{tables/method_comp_edit.tex}
% Analysis
Our approach outperforms other methods on all the evaluation metrics.
This indicates that our method produces a more accurate reconstruction with the OOD components.

\topic{Editability.}
We acquire editing directions from InterfaceGAN~\cite{shen2020interpreting} (``younger'', ``smile'') and StyleCLIP mapper~\cite{patashnik2021styleclip} (``eyeglasses'', ``surprised'', ``Elsa'').
Following previous work~\cite{roich2022pivotal,vsubrtova2022chunkygan}, we measure the ID similarity between the inverted image and the edited image, as the editing should not change a person's identity.
We report our results in Table~\ref{tab:quant_edit}. 
Our method outperforms other baselines in terms of identity preservation in most cases. 

\subsection{Qualitative results}\label{sec:qualitative}
\topic{Inversion.}
% What's shown?
We visually compare the video reconstruction in Figure~\ref{fig:qual_comp_rec}.
% What does it mean?
Our method provides higher-fidelity reconstruction results than other baselines, particularly for OOD regions (e.g., heavy make-up or earrings).
% Compared to the encoder-based method IDE-3D, our method shows more consistent results with less flickering.
Our method shows better reconstruction than the encoder-based method GOAE~\cite{Yuan_2023_GOAE} and IDE-3D~\cite{sun2022ide}. 
Compared to optimization-based methods, HFGI3D~\cite{xie2023hfgi3d}, VIVE3D~\cite{fruhstuck2023vive3d}, PTI~\cite{roich2022pivotal}, $\mathcal{W}$, and $\mathcal{W{+}}$, our method shows higher-fidelity reconstruction for OOD objects (Refer to our supplementary material for more results).

\topic{Editing.}
\input{figures/qualitative_comp_edit.tex}
% What's shown?
We show a qualitative comparison regarding the editing in Figure~\ref{fig:qual_comp_edit}.
% What does it mean?
% Our method shows faithful editing results and less temporal flickering than other baselines. 
Our method shows faithful editing results. 
For more qualitative results, please refer to our supplementary material.

% \subsection{View synthesis.}
% The use of 3D GANs supports rendering novel views after inversion. 
% We show a novel view synthesis result in Figure~\ref{fig:qual_novel_view}. 
% Our method can generate 3D consistent novel views \emph{both} for the face and OOD object.
% \input{figures/qualitative_novel_view.tex}

\subsection{Other Applications}
\topic{View synthesis.} 
The use of 3D GANs supports rendering novel views after inversion. 
We show novel view synthesis results in Figure~\ref{fig:qual_novel_view}. 
% Our method can generate 3D consistent novel views \emph{both} for the face and OOD object.
\input{figures/qualitative_novel_view.tex}

\topic{Object removal.} 
By setting the blending weights of the OOD objects to 0, we can remove OOD objects.  
We show results in Figure~\ref{fig:teaser}.
% \input{figures/qualitative_obj_removal.tex}

\input{tables/ablation.tex}
\subsection{Ablation Study}
\label{sec:ablation}
\input{figures/ablation.tex}
We introduce two new loss functions, Eqn.~\ref{eq:latent_reg} and Eqn.~\ref{eqn:blendw_loss}, to preserve the editability from the impact of the OOD radiance field in Section~\ref{sec:composite_render}. 
To validate the loss functions' effects, we conduct an ablation study in Table~\ref{tab:ablation}.
Without the weight regularization, $\mathcal{L}_b$, and latent code regularizer $\mathcal{L}_{w}$, the reconstruction accuracy is improved while the editability is reduced. 
% GAN editing is not perfect.
One of the reasons is that GAN-based editing usually also brings unwanted changes to other attributes~\cite{shen2020interpreting}. 
In Figure~\ref{fig:ablation}, the editing direction ``eyeglasses'' also moves the position of the eyes. 
% OOD weight dominate
At this time, if the blending weight $b$ is closer to 1 for pixels outside the OOD object, \ie the OOD part has more contributions, the editing tends to keep the pixel values in the reconstruction stage.
While the eyes will be moved due to the editing direction, 
it results in the duplicate eyes in Figure~\ref{fig:ablation}(a).
% Conclusion
In contrast, with regularization (Eqn.~\ref{eq:latent_reg} and Eqn.~\ref{eqn:blendw_loss}) on the blending weights, 
pixels in the in-distribution part contribute more to the output, which better supports the editing since we can only edit the in-distribution part.
Similar cases happen to $\mathcal{L}_{w}$. Without $\mathcal{L}_{w}$, the eyebrow becomes unnatural in Figure~\ref{fig:ablation} (b).

\subsection{Speed}
We include a comparison of different baselines in Table~\ref{tab:quant_inv}.
We compare the speed on 200 frames using a single NVIDIA RTX A6000 GPU.
Our method takes more time for optimization but significantly improves the reconstruction-editability trade-off. 

%% file: figures/qualitative_comp_rec.tex
\begin{figure*}
\begin{center}
\centering

$\overbrace{
\begin{subfigure}{0.11\textwidth}
    \frame{\includegraphics[trim=0 0 0 0, clip,width=\textwidth]{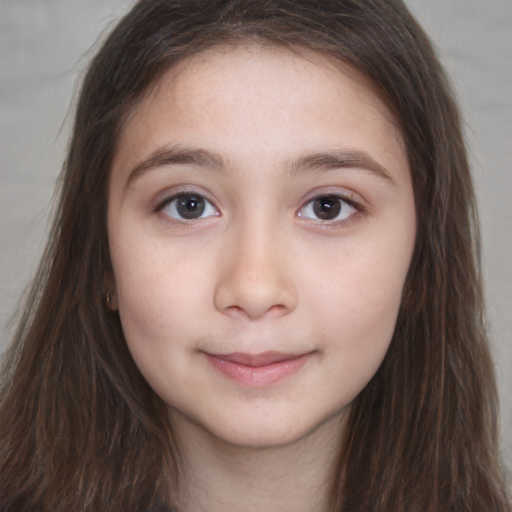}}
    \caption{\small{IDE-3D~\cite{sun2022ide}}}
\end{subfigure} 
\begin{subfigure}{0.11\textwidth}
    \frame{\includegraphics[trim=0 0 0 0, clip,width=\textwidth]{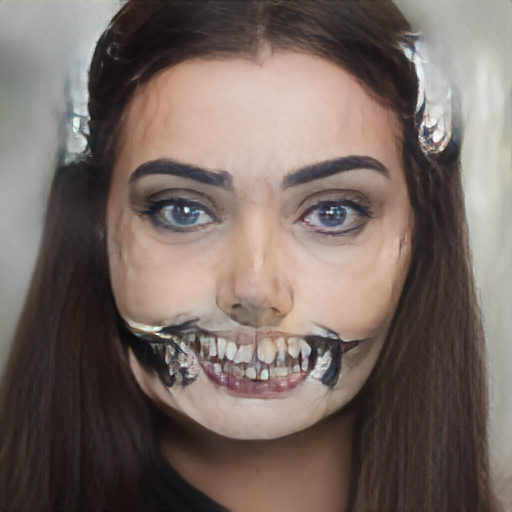}}
    \caption{\small{GOAE~\cite{Yuan_2023_GOAE}}}
\end{subfigure} 
}^{\text{Encoder-based}}$
$\overbrace{
\begin{subfigure}{0.11\textwidth}
    \frame{\includegraphics[trim=0 0 0 0, clip,width=\textwidth]{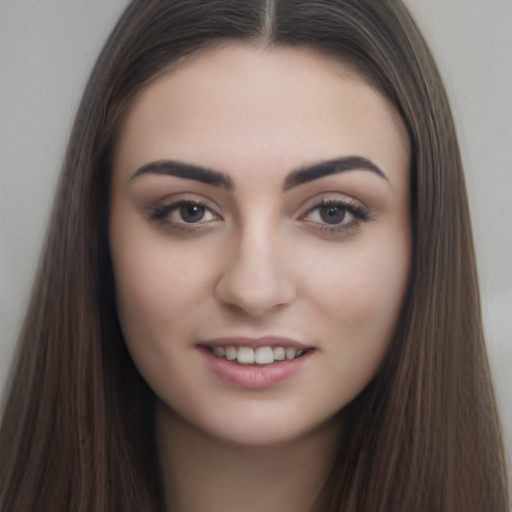}}
    \caption{\small{$\mathcal{W+}$}}
\end{subfigure}  
\begin{subfigure}{0.11\textwidth}
    \frame{\includegraphics[trim=0 0 0 0, clip,width=\textwidth]{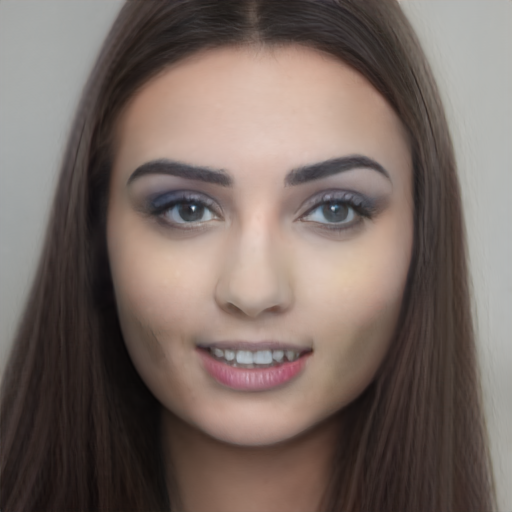}}    
    \caption{\small{$\mathcal{W}$}}
\end{subfigure}  
\begin{subfigure}{0.11\textwidth}
    \frame{\includegraphics[trim=0 0 0 0, clip,width=\textwidth]{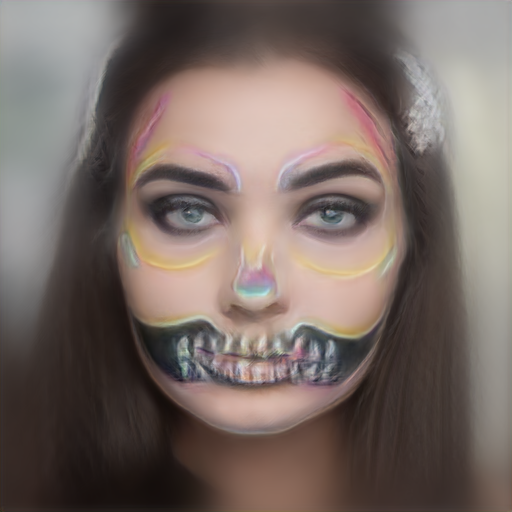}}
    \caption{\small{HFGI3D~\cite{xie2023hfgi3d}}}
\end{subfigure}  
\begin{subfigure}{0.11\textwidth}
    \frame{\includegraphics[trim=0 0 0 0, clip,width=\textwidth]{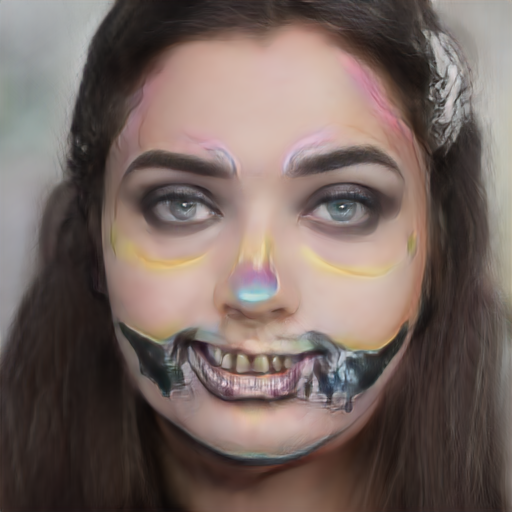}}
    \caption{\small{VIVE3D~\cite{fruhstuck2023vive3d}}}
\end{subfigure}  
\begin{subfigure}{0.11\textwidth}
    \frame{\includegraphics[trim=0 0 0 0, clip,width=\textwidth]{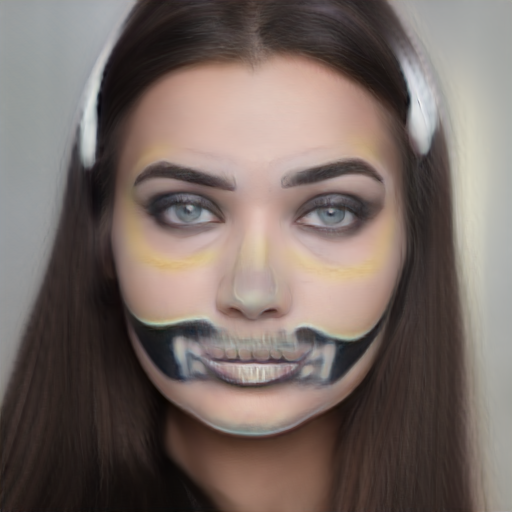}}
    \caption{\small{PTI~\cite{roich2022pivotal}}}
\end{subfigure}  
}^{\text{Optimization-based}}$
\begin{subfigure}{0.11\textwidth}
    \frame{\includegraphics[trim=0 0 0 0, clip,width=\textwidth]{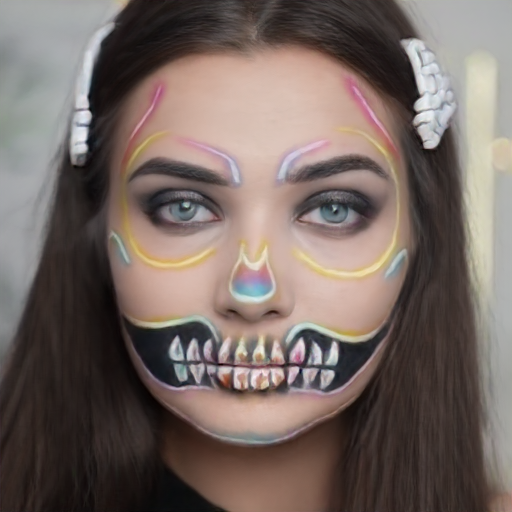}}
    \caption{\small{Ours}}
\end{subfigure} \hspace{-4pt} 
\begin{subfigure}{0.11\textwidth}
    \frame{\includegraphics[trim=0 0 0 0, clip,width=\textwidth]{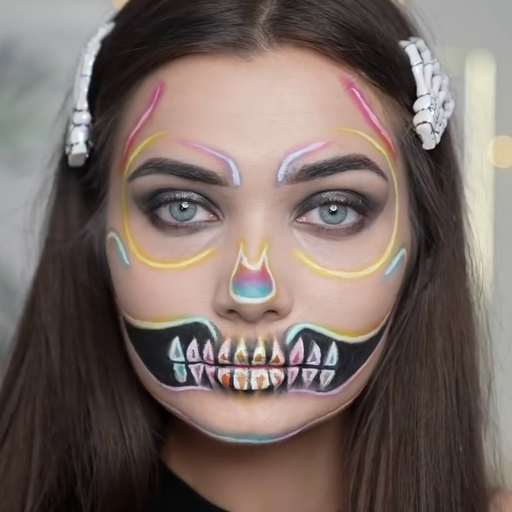}}
    \caption{\small{Input}}
\end{subfigure}\hspace{-4pt}
% \vspace{-20mm}

% \frame{\includegraphics[trim=0 0 0 0, clip,width=.11\textwidth]{images/qualitative_comp_rec/rec_comp2_wplus_00012.png}}\hspace{-20mm}\hfill
% \frame{\includegraphics[trim=0 0 0 0, clip,width=.11\textwidth]{images/qualitative_comp_rec/rec_comp2_w_00012.png}}\hspace{-20mm}\hfill
% \frame{\includegraphics[trim=0 0 0 0, clip,width=.11\textwidth]{images/qualitative_comp_rec/rec_comp2_ide3d_00012.png}}\hspace{-20mm}\hfill
% \frame{\includegraphics[trim=0 0 0 0, clip,width=.11\textwidth]{images/qualitative_comp_rec/rec_comp2_pti_00012.png}}\hspace{-20mm}\hfill
% \frame{\includegraphics[trim=0 0 0 0, clip,width=.11\textwidth]{images/qualitative_comp_rec/rec_comp2_ours_00012.png}}\hspace{-20mm}\hfill
% \frame{\includegraphics[trim=0 0 0 0, clip,width=.11\textwidth]{images/qualitative_comp_rec/rec_comp2_target_00012.png}}\hspace{-20mm}\\

% \mpage{0.1}{\small{IDE-3D~\cite{sun2022ide}}}\hfill
% \mpage{0.1}{\small{GOAE~\cite{Yuan_2023_GOAE}}}\hfill
% \mpage{0.1}{\small{$\mathcal{W}^{+}$}}\hfill
% \mpage{0.1}{\small{$\mathcal{W}$}}\hfill
% \mpage{0.1}{\small{HFGI3D~\cite{xie2023hfgi3d}}}\hfill
% \mpage{0.1}{\small{VIVE3D~\cite{fruhstuck2023vive3d}}}\hfill
% \mpage{0.1}{\small{PTI~\cite{roich2022pivotal}}}\hfill
% \mpage{0.1}{\small{Ours}}\hfill
% \mpage{0.1}{\small{Input}}\\

\caption{
\textbf{Qualitative comparison of the video reconstruction.} 
We compare our approach with $\mathcal{W+}$ and $\mathcal{W}$ optimization, IDE-3D~\cite{sun2022ide}, GOAE~\cite{Yuan_2023_GOAE}, HFGI3D~\cite{xie2023hfgi3d}, VIVE3D~\cite{fruhstuck2023vive3d}, and PTI~\cite{roich2022pivotal}. Our method shows a better reconstruction accuracy on the OOD videos. 
}
\label{fig:qual_comp_rec}
\end{center}
\vspace{-6mm}
\end{figure*}

%% file: tables/method_comp_inv.tex
\definecolor{tabfirst}{rgb}{1, 0.7, 0.7} % red
\definecolor{tabsecond}{rgb}{1, 0.85, 0.7} % orange
\definecolor{tabthird}{rgb}{1, 1, 0.7} % yellow

\begin{table*}[t!]
  \centering
  \scriptsize
  
    \begin{tabular}{lrrrr|rrrrr}
    \toprule
          & \multicolumn{4}{c|}{Images}           & \multicolumn{5}{c}{Videos} \\
\cmidrule{1-10}          & \multicolumn{1}{c}{LPIPS$\downarrow$}  & \multicolumn{1}{c}{SSIM$\uparrow$} & \multicolumn{1}{c}{PSNR$\uparrow$} & \multicolumn{1}{c|}{ID Similarity$\uparrow$} & \multicolumn{1}{c}{Time$\downarrow$} & \multicolumn{1}{c}{LPIPS$\downarrow$}  & \multicolumn{1}{c}{SSIM$\uparrow$} & \multicolumn{1}{c}{PSNR$\uparrow$} & \multicolumn{1}{c}{ID Similarity$\uparrow$} \\
    \midrule
    Ours                        &  \cellcolor{tabfirst}0.1106 &  \cellcolor{tabfirst}0.8175 &  \cellcolor{tabfirst}19.86 &  \cellcolor{tabfirst}0.9685   & 2.68h  &     \cellcolor{tabfirst}0.2237 &  \cellcolor{tabfirst}0.7052 &  \cellcolor{tabfirst}16.03 &  \cellcolor{tabfirst}0.9758 \\
    HFGI3D~\cite{xie2023hfgi3d} & 0.3912 & 0.5521 & 11.37 &  \cellcolor{tabthird}0.9463    & 7.51h &   0.3954 & 0.5587 & 11.55 &  \cellcolor{tabthird}0.9388 \\
    GOAE~\cite{Yuan_2023_GOAE}  & 0.3619 & 0.6424 & 14.73 &  \cellcolor{tabfirst}0.9685 &  {56s}  &  0.3642 &  \cellcolor{tabthird}0.6470 & \cellcolor{tabsecond}14.97 & 0.3642 \\
    E3DGE~\cite{lan2023e3dge}& \cellcolor{tabsecond}0.1709 & \cellcolor{tabsecond}0.7738 & \cellcolor{tabsecond}15.28 &  0.8632    & - &   - & - & - &  - \\
    VIVE3D~\cite{fruhstuck2023vive3d} &  -     &  -     &    -   &  -    &  0.59h  &  0.4172 & 0.5417 & 10.66 & 0.9245 \\
    PTI~\cite{roich2022pivotal} & 0.3192 & 0.6172 & 12.93 & \cellcolor{tabsecond}0.9676  & 1.45h   &  \cellcolor{tabsecond}0.3144 & 0.6320 & 13.45 & \cellcolor{tabsecond}0.9658 \\
    IDE-3D~\cite{sun2022ide}    & 0.5044 & 0.4395 & 9.18 & 0.8456  &  77s  &  0.4999 & 0.4512 & 9.59 & 0.8251 \\
   $\mathcal{W{+}}$           &  \cellcolor{tabthird}0.3433 &  \cellcolor{tabthird}0.6387 &  \cellcolor{tabthird}14.39 & 0.9199   &  0.49h  &   \cellcolor{tabthird}0.3380 & \cellcolor{tabsecond}0.6557 &  \cellcolor{tabthird}14.75 & 0.9154 \\
    $\mathcal{W}$               & 0.4097 & 0.5615 & 12.08 & 0.8757  &  0.47h  & 0.4030 & 0.5787 & 12.48 & 0.8652 \\
    \bottomrule
    \end{tabular}%
\caption{\textbf{Reconstruction quality evaluation}. For each column, deeper color the better. 
      % The speed is tested on a 200-frame video with a single RTX A6000 GPU.
  }
  \label{tab:quant_inv}%
\end{table*}%

%% file: tables/method_comp_edit.tex
\begin{table*}[t!]
  \centering
  \scriptsize
    \begin{tabular}{lrrrrrr|rrrrrr}
    \toprule
          & \multicolumn{6}{c|}{Images}                   & \multicolumn{6}{c}{Videos} \\
\cmidrule{1-13}          & \multicolumn{1}{c}{eyeglasses} & \multicolumn{1}{c}{surprised} & \multicolumn{1}{c}{younger} & \multicolumn{1}{c}{smile} & \multicolumn{1}{c}{Elsa} & \multicolumn{1}{c|}{average} & \multicolumn{1}{c}{eyeglasses} & \multicolumn{1}{c}{surprised} & \multicolumn{1}{c}{younger} & \multicolumn{1}{c}{smile} & \multicolumn{1}{c}{Elsa} & \multicolumn{1}{c}{average} \\
    \midrule
    Ours                        &  \cellcolor{tabfirst}.9532 &  \cellcolor{tabfirst}.9888 &  \cellcolor{tabfirst}.9495 &  \cellcolor{tabfirst}.9525 &  \cellcolor{tabfirst}.9116 &  \cellcolor{tabfirst}.9511    &    \cellcolor{tabfirst}.9158 & .9360 &  \cellcolor{tabfirst}.9347 & .9094 &  \cellcolor{tabfirst}.8927 &  \cellcolor{tabfirst}.9177 \\
    HFGI3D~\cite{xie2023hfgi3d} & \cellcolor{tabsecond}.9484 & \cellcolor{tabsecond}.9795 & \cellcolor{tabsecond}.9453 & .9223 & .8641 & \cellcolor{tabsecond}.9319     &  \cellcolor{tabthird}.9112 & .9109 &  \cellcolor{tabthird}.9290 & .9155 & .8622 & .9058 \\
    GOAE~\cite{Yuan_2023_GOAE}  &  \cellcolor{tabthird}.9179 & .9306 & .9327 & .9332 & \cellcolor{tabsecond}.8851 &  \cellcolor{tabthird}.9199     &         \cellcolor{tabsecond}.9120 & .9224 & .9235 &  \cellcolor{tabthird}.9221 &  \cellcolor{tabthird}.8641 &  \cellcolor{tabthird}.9088 \\
    E3DGE~\cite{lan2023e3dge}&  - & - & .8853 & \cellcolor{tabsecond}.9487 & - &  0.9170     &  - & - & - &  - &  - &  - \\
    VIVE3D~\cite{fruhstuck2023vive3d} &  -     &    -   &  -     &    -   &  - & -  &  .9078 &  \cellcolor{tabthird}.9475 & .9183 &  \cellcolor{tabfirst}.9369 & \cellcolor{tabsecond}.8728 & \cellcolor{tabsecond}.9167 \\
    PTI~\cite{roich2022pivotal} & .9114 & .9562 &  \cellcolor{tabthird}.9380 & .9410 & .7927 & .9079     &  .9049 & .9357 & \cellcolor{tabsecond}.9319 & \cellcolor{tabsecond}.9336 & .7945 & .9001 \\
    IDE-3D~\cite{sun2022ide}    & .8811 & .9538 & .8723 & .8055 &  \cellcolor{tabthird}.8780 & .8781     &  .8767 & \cellcolor{tabsecond}.9481 & .8551 & .8662 & .7871 & .8666 \\
    $\mathcal{W{+}}$           & .9012 &  \cellcolor{tabthird}.9567 & .9248 &  \cellcolor{tabthird}.9356 & .7892 & .9015     &  .8971 & .9249 &  \cellcolor{tabthird}.9290 & .9170 & .7968 & .8930 \\
    $\mathcal{W}$               & .8808 &  \cellcolor{tabthird}.9567 & .9177 & .9290 & .8008 & .8970   &  .8793 &  \cellcolor{tabfirst}.9537 & .9068 & .9208 & .8113 & .8944 \\
    \bottomrule
    \end{tabular}%
  \caption{\textbf{Identity preservation evaluation.} Higher numbers indicate better identity preservation. $\mathcal{W{+}}$ is equivalent to our method without OOD tri-plane.}
  \vspace{-5mm}
  \label{tab:quant_edit}%
\end{table*}%

% \begin{table}
%     \centering
%     \begin{tabular}{l|ccc}
%     & PSNR $\uparrow$ & LPIPS $\downarrow$ & Runtime \\ \hline

%     That one algorithm    &  \cellcolor{tabthird}3.52 &  \cellcolor{tabthird}.151 & 10 sec. \\
% That other algorithm  & \cellcolor{tabsecond}32.13 & \cellcolor{tabsecond}.074 & 100 sec. \\
% Yet another algorithm &                      19.26 &                      .433 & 3 sec. \\
% \hline
% My beloved algorithm  &  \cellcolor{tabfirst}38.92 &  \cellcolor{tabfirst}.051 & 60 sec.

%     \end{tabular}
%     \caption{Wow what great results}
%     \label{tab:results}
% \end{table}

%% file: figures/qualitative_comp_edit.tex
\begin{figure*}
\begin{center}
\centering

\mpage{0.01}{\raisebox{50pt}{\rotatebox{90}{Image}}} 
\frame{\includegraphics[trim=0 0 0 0, clip,width=.11\textwidth]{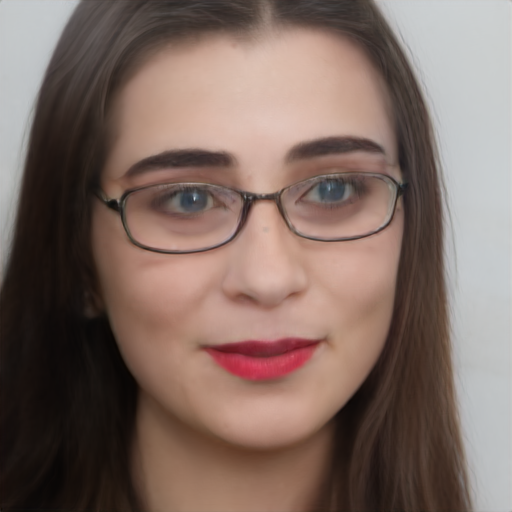}}\hspace{-20mm}\hfill
\frame{\includegraphics[trim=0 0 0 0, clip,width=.11\textwidth]{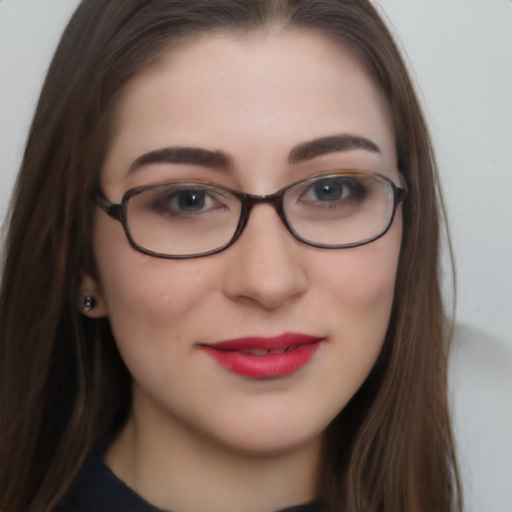}}\hspace{-20mm}\hfill
\frame{\includegraphics[trim=0 0 0 0, clip,width=.11\textwidth]{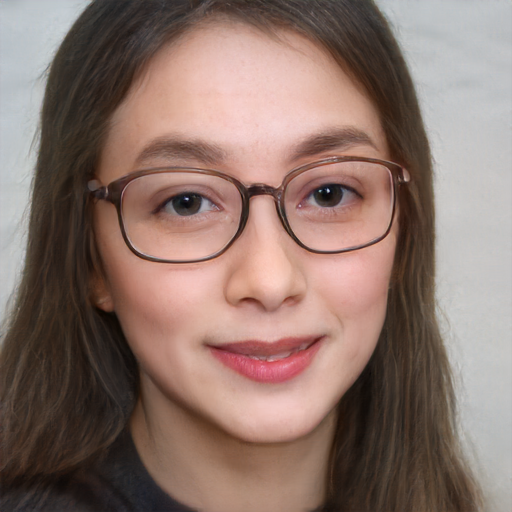}}\hspace{-20mm}\hfill
\frame{\includegraphics[trim=0 0 0 0, clip,width=.11\textwidth]{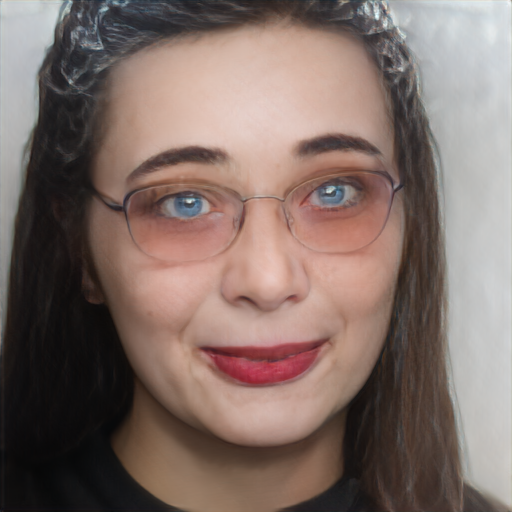}}\hspace{-20mm}\hfill
\frame{\includegraphics[trim=0 0 0 0, clip,width=.11\textwidth]{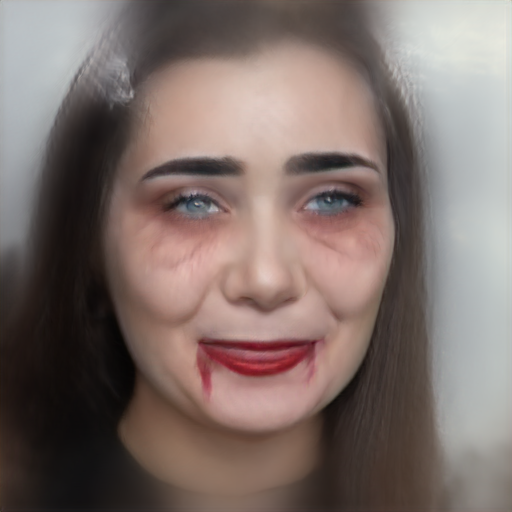}}\hspace{-20mm}\hfill
\frame{\includegraphics[trim=0 0 0 0, clip,width=.11\textwidth]{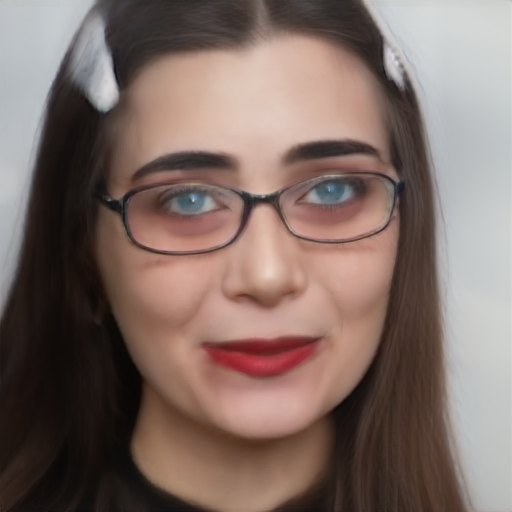}}\hspace{-20mm}\hfill
\frame{\includegraphics[trim=0 0 0 0, clip,width=.11\textwidth]{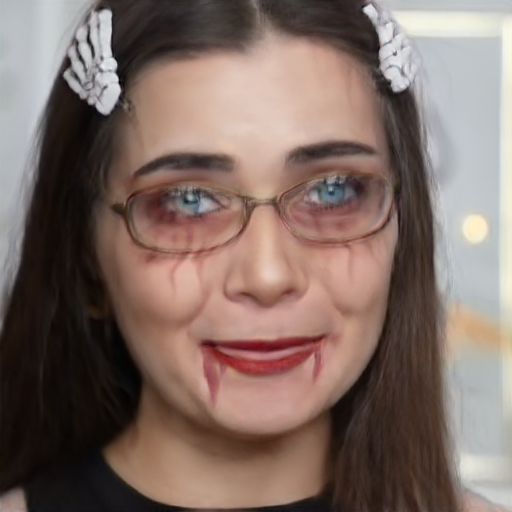}}\hspace{-20mm}\hfill
\frame{\includegraphics[trim=0 0 0 0, clip,width=.11\textwidth]{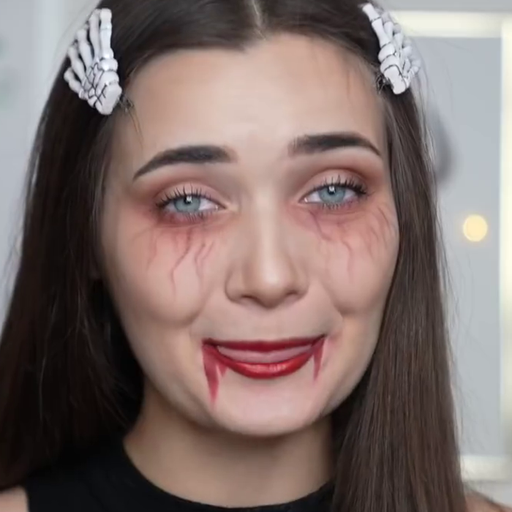}}\hspace{-20mm}\\

\vspace{-12mm}
\mpage{0.12}{\small{$\mathcal{W}^{+}$}}\hfill
\mpage{0.1}{\small{$\mathcal{W}$}}\hfill
\mpage{0.1}{\small{IDE-3D~\cite{sun2022ide}}}\hfill
\mpage{0.1}{\small{GOAE~\cite{Yuan_2023_GOAE}}}\hfill
\mpage{0.1}{\small{HFGI3D~\cite{xie2023hfgi3d}}}\hfill
\mpage{0.1}{\small{PTI~\cite{roich2022pivotal}}}\hfill
\mpage{0.1}{\small{Ours}}\hfill
\mpage{0.1}{\small{Input}}\\

\mpage{0.01}{\raisebox{50pt}{\rotatebox{90}{Video}}} 
\frame{\includegraphics[trim=0 0 0 0, clip,width=.10\textwidth]{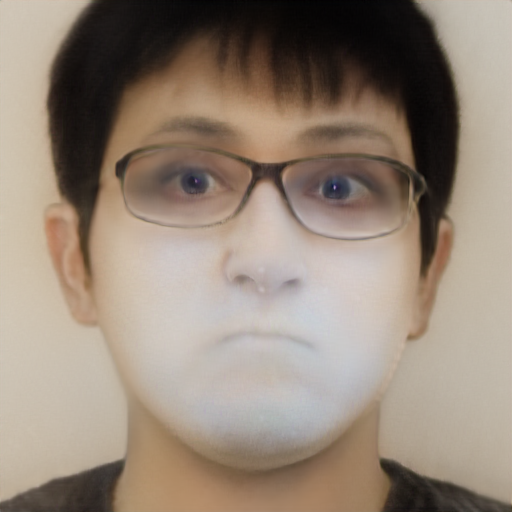}}\hspace{-20mm}\hfill
\frame{\includegraphics[trim=0 0 0 0, clip,width=.10\textwidth]{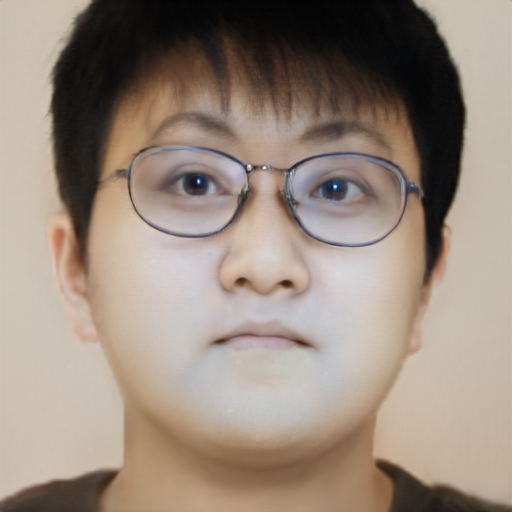}}\hspace{-20mm}\hfill
\frame{\includegraphics[trim=0 0 0 0, clip,width=.10\textwidth]{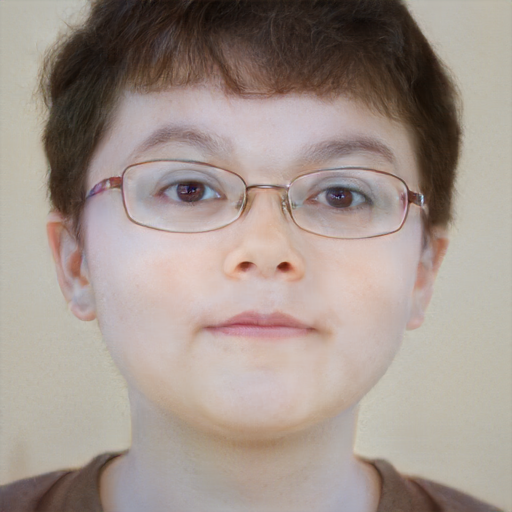}}\hspace{-20mm}\hfill
\frame{\includegraphics[trim=0 0 0 0, clip,width=.10\textwidth]{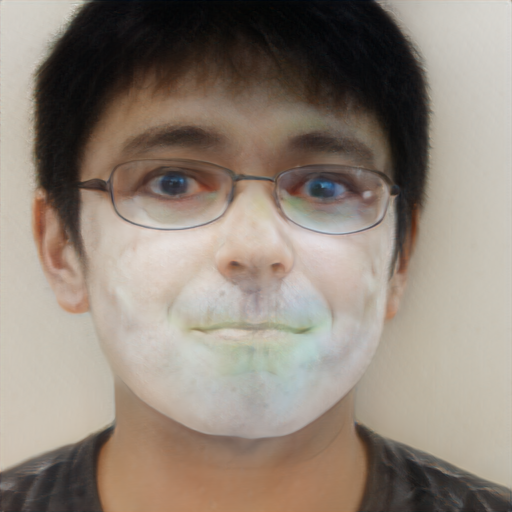}}\hspace{-20mm}\hfill
\frame{\includegraphics[trim=0 0 0 0, clip,width=.10\textwidth]{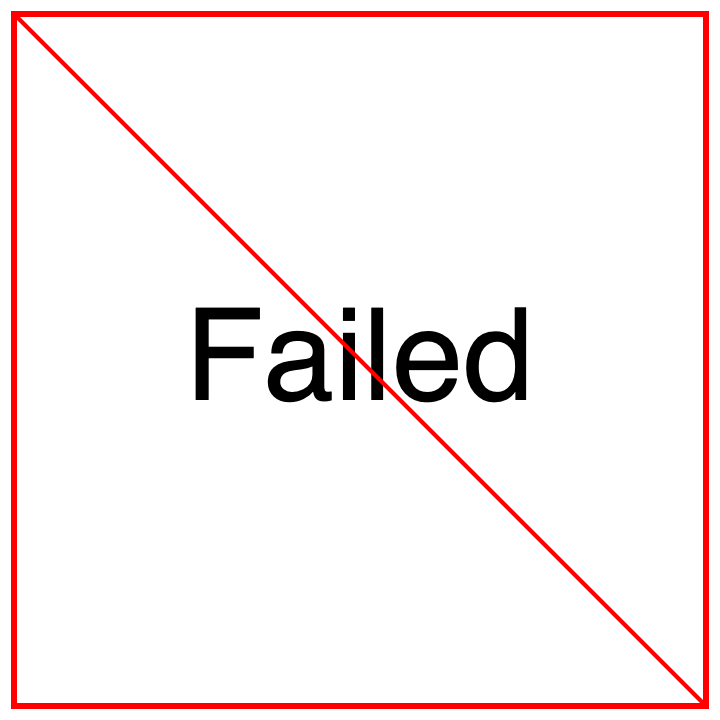}}\hspace{-20mm}\hfill
\frame{\includegraphics[trim=0 0 0 0, clip,width=.10\textwidth]{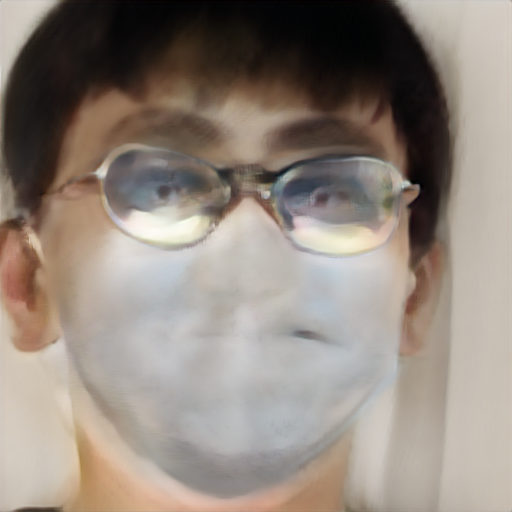}}\hspace{-20mm}\hfill
\frame{\includegraphics[trim=0 0 0 0, clip,width=.10\textwidth]{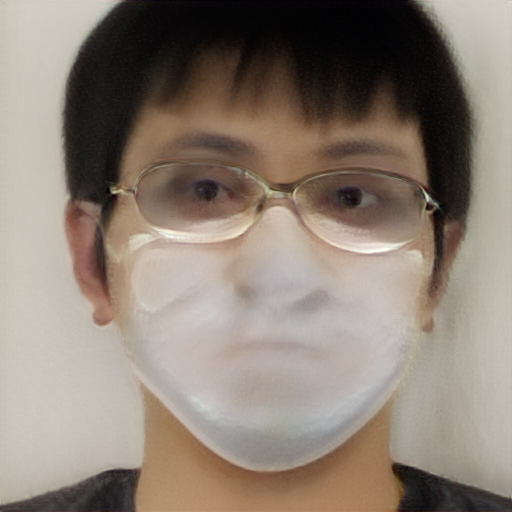}}\hspace{-20mm}\hfill
\frame{\includegraphics[trim=0 0 0 0, clip,width=.10\textwidth]{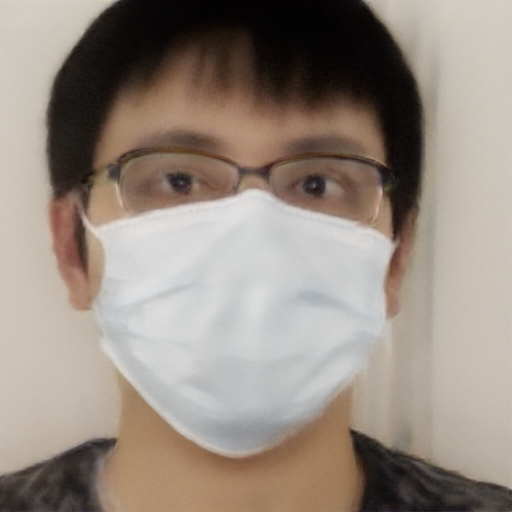}}\hspace{-20mm}\hfill
\frame{\includegraphics[trim=0 0 0 0, clip,width=.10\textwidth]{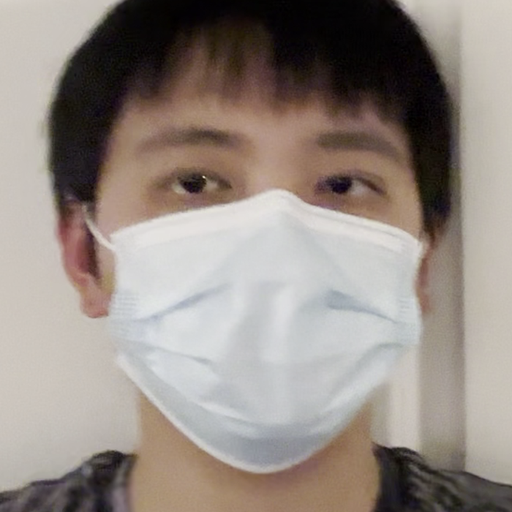}}\hspace{-20mm}\\

% \frame{\includegraphics[trim=0 0 0 0, clip,width=.11\textwidth]{images/qualitative_comp_edit/Halloween6_wplus_surprised_00000.png}}\hspace{-20mm}\hfill
% \frame{\includegraphics[trim=0 0 0 0, clip,width=.11\textwidth]{images/qualitative_comp_edit/Halloween6_w_surprised_00000.png}}\hspace{-20mm}\hfill
% \frame{\includegraphics[trim=0 0 0 0, clip,width=.11\textwidth]{images/qualitative_comp_edit/Halloween6_ide3d_surprised_00000.png}}\hspace{-20mm}\hfill
% \frame{\includegraphics[trim=0 0 0 0, clip,width=.11\textwidth]{images/qualitative_comp_edit/Halloween6_goae_surprised_00000.png}}\hspace{-20mm}\hfill
% \frame{\includegraphics[trim=0 0 0 0, clip,width=.11\textwidth]{images/qualitative_comp_edit/Halloween6_hfgi3d_surprised_00000.png}}\hspace{-20mm}\hfill
% \frame{\includegraphics[trim=0 0 0 0, clip,width=.11\textwidth]{images/qualitative_comp_edit/Halloween6_vive3d_surprised_00000.png}}\hspace{-20mm}\hfill
% \frame{\includegraphics[trim=0 0 0 0, clip,width=.11\textwidth]{images/qualitative_comp_edit/Halloween6_pti_surprised_00000.png}}\hspace{-20mm}\hfill
% \frame{\includegraphics[trim=0 0 0 0, clip,width=.11\textwidth]{images/qualitative_comp_edit/Halloween6_ours_surprised_00000.png}}\hspace{-20mm}\hfill
% \frame{\includegraphics[trim=0 0 0 0, clip,width=.11\textwidth]{images/qualitative_comp_edit/Halloween6_target_00000.png}}\hspace{-20mm}\\

\vspace{-12mm}
\mpage{0.12}{\small{$\mathcal{W}^{+}$}}\hfill
\mpage{0.1}{\small{$\mathcal{W}$}}\hfill
\mpage{0.1}{\small{IDE-3D~\cite{sun2022ide}}}\hfill
\mpage{0.1}{\small{GOAE~\cite{Yuan_2023_GOAE}}}\hfill
\mpage{0.1}{\small{HFGI3D~\cite{xie2023hfgi3d}}}\hfill
\mpage{0.1}{\small{VIVE3D~\cite{fruhstuck2023vive3d}}}\hfill
\mpage{0.1}{\small{PTI~\cite{roich2022pivotal}}}\hfill
\mpage{0.1}{\small{Ours}}\hfill
\mpage{0.1}{\small{Input}}\\
\vspace{-3mm}
\caption{
\textbf{Qualitative comparison of the editing.} 
We compare our editing results from a single image and a video with other baselines, with different editing latent directions ``Eyeglasses''.
Our approach can preserve the original appearance details better,  and shows improved editability over other baselines. 
% Videos are from Internet (Creative Commons). 
}
\label{fig:qual_comp_edit}
\end{center}
\vspace{-8mm}
\end{figure*}

%% file: figures/qualitative_novel_view.tex
\begin{figure}
\begin{center}
\centering

\frame{\includegraphics[trim=0 0 0 0, clip,width=.150\textwidth]{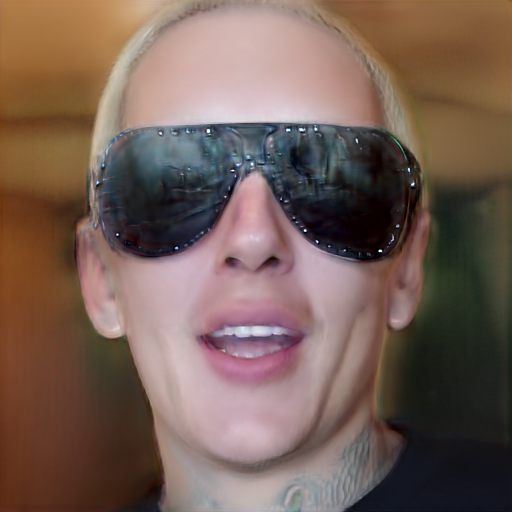}}\hspace{-20mm}\hfill
\frame{\includegraphics[trim=0 0 0 0, clip,width=.150\textwidth]{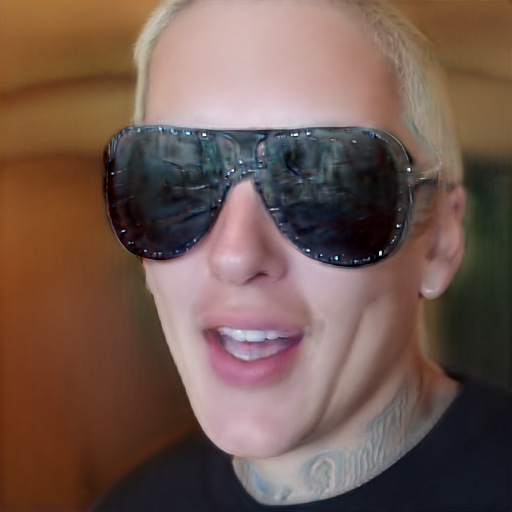}}\hspace{-20mm}\hfill
\frame{\includegraphics[trim=0 0 0 0, clip,width=.150\textwidth]{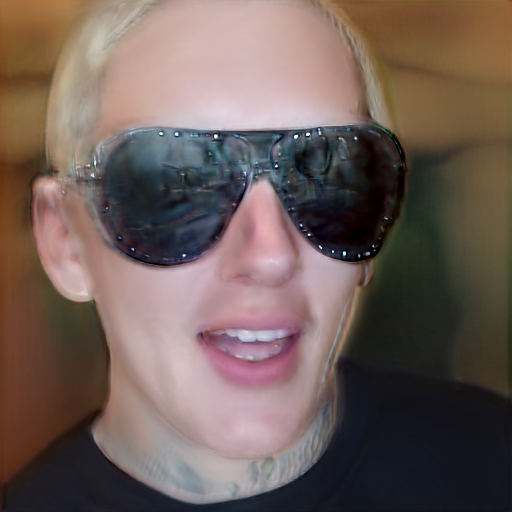}}\hspace{-20mm}\\

% \vspace{2mm}

% \frame{\includegraphics[trim=0 0 0 0, clip,width=.150\textwidth]{images/novel_view_synthesis/Halloween3_nv_frame0000.png}}\hspace{-20mm}\hfill
% \frame{\includegraphics[trim=0 0 0 0, clip,width=.150\textwidth]{images/novel_view_synthesis/Halloween3_nv_frame0020.png}}\hspace{-20mm}\hfill
% \frame{\includegraphics[trim=0 0 0 0, clip,width=.150\textwidth]{images/novel_view_synthesis/Halloween3_nv_frame0095.png}}\\

\caption{
\textbf{Novel view synthesis.} 
We can synthesize novel views for a fixed frame in a video, which is challenging for 2D GANs. Each column shows different view for the same frame. 
}
\label{fig:qual_novel_view}
\end{center}
\vspace{-5mm}
\end{figure}

%% file: tables/ablation.tex
% \vspace{-5mm}
\begin{table}[h!]
  \centering
  \scriptsize
    \begin{tabular}{l|rr|r}
    \toprule
          & \multicolumn{2}{c|}{Inversion} & \multicolumn{1}{c}{Editing} \\
    \midrule
          & \multicolumn{1}{l}{$\mathcal{L}_2\downarrow$} & \multicolumn{1}{l|}{LPIPS$\downarrow$} & \multicolumn{1}{l}{ID similarity$\uparrow$} \\
    \midrule
    % w/o $\mathcal{L}_b$ and $\mathcal{L}_{w}$ &    0.0362   &  0.2339      & 0.8826   \\
    w/o $\mathcal{L}_b$   &  \cellcolor{tabfirst}0.0322 &  \cellcolor{tabfirst}0.2191 & \cellcolor{tabsecond}0.9070 \\
    w/o $\mathcal{L}_{w}$ & \cellcolor{tabsecond}0.0336 &  \cellcolor{tabthird}0.2238 &  \cellcolor{tabthird}0.9024 \\
    Full method           &  \cellcolor{tabthird}0.0339 & \cellcolor{tabsecond}0.2237 &  \cellcolor{tabfirst}0.9177 \\
    \bottomrule
    \end{tabular}%
  \caption{\textbf{Ablation study.}
  We study the effect of different loss functions on 20 videos. 
  For inversion, we compute the metrics between reconstructed frames and input frames.
  For editing, we compute the ID similarity between before and after editing. 
  }
  \vspace{-5mm}
  \label{tab:ablation}%
\end{table}%

%% file: figures/ablation.tex
\begin{figure}
\begin{center}
\centering

% \mpage{0.01}{\raisebox{60pt}{{}}} \hfill
% \frame{\includegraphics[trim=0 0 0 0, clip,width=.11\textwidth]{images/ablation/noBlendw_LatentReg_proj_00000.png}}\hfill
\frame{\includegraphics[trim=0 0 0 0, clip,width=.11\textwidth]{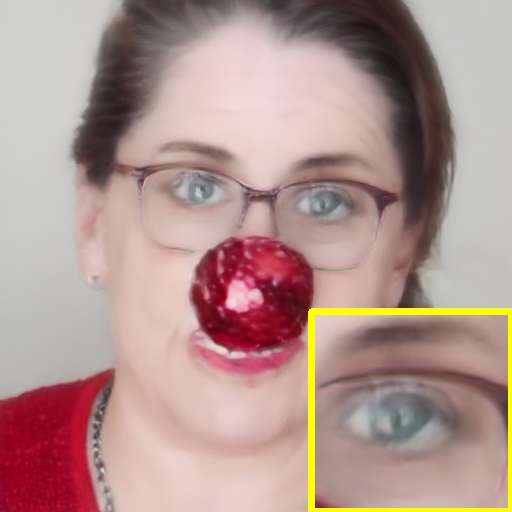}}\hspace{-5pt}\hfill
\frame{\includegraphics[trim=0 0 0 0, clip,width=.11\textwidth]{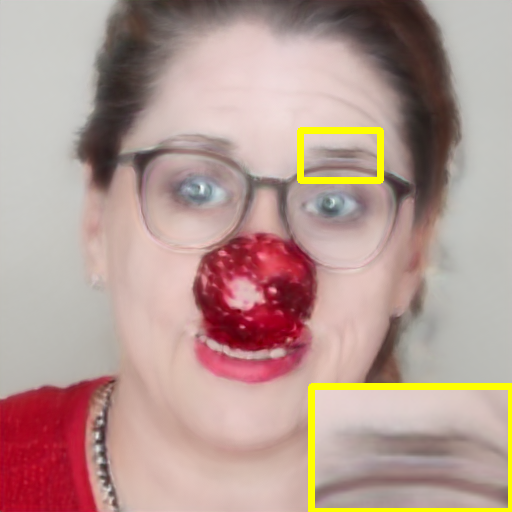}}\hspace{-5pt}\hfill
% \frame{\includegraphics[trim=0 0 0 0, clip,width=.11\textwidth]{images/ablation/LargeLr_proj_00000.png}}\hfill
\frame{\includegraphics[trim=0 0 0 0, clip,width=.11\textwidth]
{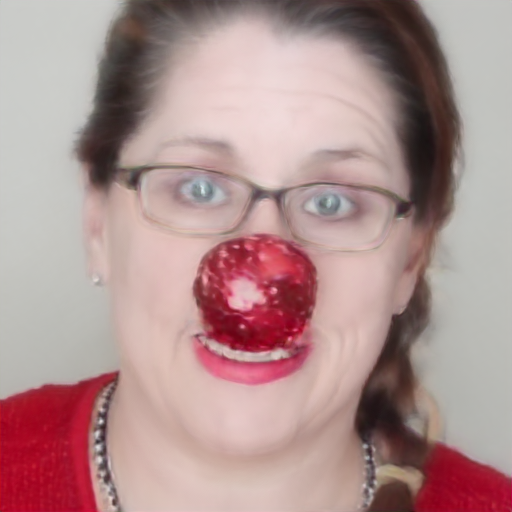}}\\

% \mpage{0.2}{\small{w/o $\mathcal{L}_b$ and $\mathcal{L}_{w}$}}\hfill
\mpage{0.2}{\small{(a) w/o $\mathcal{L}_b$}}\hfill
\mpage{0.2}{\small{(b) w/o $\mathcal{L}_{w}$}}\hfill
% \mpage{0.2}{\small{Large learning rate}}\hfill
\mpage{0.25}{\small{(c) Full model}}

\caption{
\textbf{Ablation study on editing.} 
(a) Without $\mathcal{L}_b$, the out-of-distribution component dominates ($b \rightarrow 1$) and weakens the editing. It has
``duplicate eyes'' artifact because the editing direction ``eyeglasses'' is not disentangled well with other attributes, and changes the positions of the eyes, while the blending weights are the same as the reconstruction,
it results in duplicated eyes.
(b) Without $\mathcal{L}_w$, the eyebrow becomes unnatural. 
}
\label{fig:ablation}
\end{center}
\vspace{-8mm}
\end{figure}

%% file: 5_limitations.tex
\section{Limitations}
\label{sec:limitations}
% Our approach consists of the following limitations. 
\input{figures/limitations.tex}
Our method still has several limitations.
We visualize (a)-(c) in Figure~\ref{fig:limitations}.

\topic{(a) Editing on OOD part.} When editing on the OOD region, \eg adding eyeglasses to the heavy makeup region, because the blending weights are closer to 1, the eyeglasses in the in-distribution radiance field are hard to be added. 

\topic{(b) Duplicate objects.} Since our OOD radiance field has no knowledge about the GAN and faces prior, when the OOD object itself is glasses, adding eyeglasses introduces duplicate objects. 

\topic{(c) Extreme poses.} Our method fails at editing when the subject undergoes extreme poses (e.g., side view).

\topic{(d) Objects with limited movement.} The radiance field reconstruction suffers when the OOD object has slight movement. This may introduce unwanted artifacts like ``floater'' in the novel views. 

\topic{(e) Temporal inconsistency.} 
Our results on video editing may suffer from temporal inconsistency. 
Temporal constraints and finetuning used in \cite{xu2022temporally,tzaban2022stitch} could further improve this aspect.
% We do not have temporal constraints for videos, thus our model may introduce temporal inconsistency.

% \topic{Novel view synthesis from single image.} Since our method needs multi-view input, it is challenging for our method to render novel view if only a single image input is given. 

% \topic{Temporal consistency.} Our method lacks constraints in the temporal axis. 
% It is also known that, even if the inversion result is enough temporally consistent, the editing result can still be temporally flickering~\cite{xu2022temporally}.

%% file: figures/limitations.tex
\begin{figure}
\begin{center}
\centering

\frame{\includegraphics[trim=0 0 0 0, clip,width=.13\textwidth]{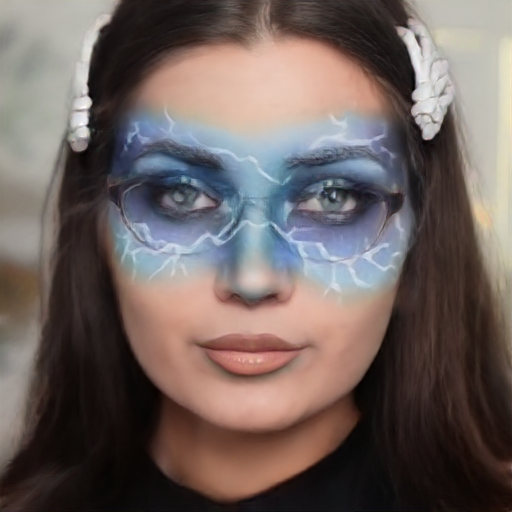}}\hfill
\frame{\includegraphics[trim=0 0 0 0, clip,width=.13\textwidth]{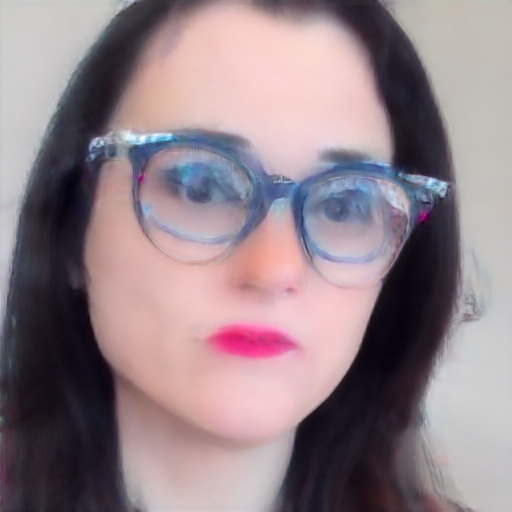}}\hfill
\frame{\includegraphics[trim=0 0 0 0, clip,width=.13\textwidth]{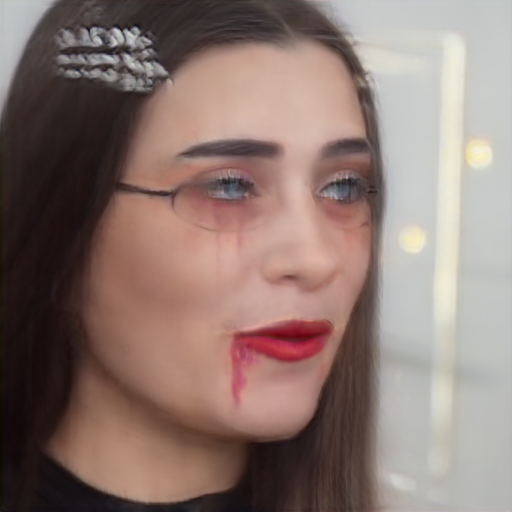}}\\

\mpage{0.3}{\small{(a) OOD dominates}}\hfill
\mpage{0.3}{\small{(b) Double glasses}}\hfill
\mpage{0.3}{\small{(c) Extreme pose}}\\

\caption{
\textbf{Limitations.} 
Our approach has some limitations.
(a) Editing on where OOD blending weights dominate is challenging,
(b) Adding another eyeglasses to OOD eyeglasses will result in duplicated objects, and (c) extreme poses.
}
\label{fig:limitations}
\end{center}
\vspace{-8mm}
\end{figure}

%% file: 6_conclusions.tex
\section{Conclusions}
\label{sec:conclusions}
% What we have done
We have presented a novel method for face image, and its potential for video inversion and editing. 
Our method handles OOD objects by isolating them from the InD part.
Our method achieves accurate reconstruction by building two radiance fields and then composing them together during the rendering.
By modifying the latent code in the InD part, we can obtain faithful editing results.
% How is the result?
We show that our method achieves a better balance in the reconstruction-editability trade-off than other baselines.
% \topic{Social impacts.} 
Malicious use of our technique may lead to misinformation.

%% file: supp_content.tex
In this supplementary material, we present additional visual results and implementation details to complement the main paper.

\section{Overview}
We include the following contents as our supplementary material:
\begin{itemize}
    \item The implementation details of our proposed approach. Our code and models will be publicly available upon publication.
    \item Additional results, organized in our project page - \href{https://in-n-out-3d.github.io/}{https://in-n-out-3d.github.io/}. 
\end{itemize}

\section{Implementation details}
This section discusses further details not included in the main paper due to the page limit.

\subsection{In-distribution inversion}
We further present the details in Section {\color{blue}4.1} in the main paper. 

\topic{Latent code initialization.} 
Instead of initializing the latent codes $[w_1, \cdots, w_N], w_t \in \mathbb{R}^{14 \times 512}$ using an average latent code, we propose to use the nearest neighbor (NN) approach for the initialization. 
We show our approach in Algorithm~\ref{alg:nn_init}. Please recall that we use $N=1$ for single image.

\input{algorithms/NN_initial.tex}

\topic{Optimization.} 
We use a residual form to represent the latent code of each frame.
\begin{equation}
    w_t = w^{cano} + a w^{res}_t \,,
\end{equation}
where $w^{cano}$ is the canonical latent code, and $w^{res}_t$ is the residual. In practice, we use $a = 0.7$.

% During the optimization, we first sample $K=16$ frames from the entire video.
% For the subset optimization, we use $M_1 = 200$; to upscale the result to the whole video, we use $M_2 = 20$. 
% The optimizer is Adam~\cite{kingma2014adam}. 
% The learning rate is $1\times10^{-3}$.
To initialize the latent codes, we randomly sample $n=500$ latent codes.

\subsection{Out-of-domain inversion}
We further explain the details in Section {\color{blue}4.2} in the main paper. 

\topic{Tri-plane.}
We use the Tri-plane representation from EG3D~\cite{chan2022efficient}. 
For the out-of-distribution (OOD) object, we use tri-plane $\mathbf{T}^{O} \in \mathbb{R}^{256\times256\times32\times3}$ as its 3D representation. 
$\mathbf{T}^{O}$ is initialized from a standard normal distribution. 
Given a 3D position $\mathbf{x} \in \mathbb{R}^{3}$, we project it onto each of the plane and retrieve the corresponding feature vectors $F_{xy}, F_{yz}, F_{xz}$ through a bilinear interpolation. 
Then three feature vectors are aggregated via summation. 

\topic{Per-frame latent code $\mathbf{\phi}_t$.} To encode the OOD object across different frames, we use a time-varying latent code $\mathbf{\phi}_t \in \mathbb{R}^{32}$ for each frame. We draw $\mathbf{\phi}_t \sim \mathcal{N}(\mathbf{0}, \mathbf{I})$. 
$\mathbf{\phi}_t$ is optimized together with other variables in Section {\color{blue}4.2}.

\topic{Decoder.} For the out-of-distribution inversion, we use a two-layer Multi-Layer Perceptron (MLP) as the decoder to predict the pointwise color $\mathbf{c} \in \mathbb{R}^3$, density $\sigma \in \mathbb{R}$, and blending weight $b \in \mathbb{R}$ for the later composite rendering.
\begin{equation}
    (\mathbf{c}^O, \sigma^O, b) = D^O(\mathbf{T}^O(t_k), \mathbf{\phi}_t; \theta_{D^O}) \,.
\end{equation}

\topic{Optimization.} During the optimization, we use the loss function Eqn. {\color{blue}8} in the main paper. 

\subsection{Overall optimization}
In practice, we only optimize with the total loss function (Eqn.  {\color{blue}8} in main paper, also shown below) for both in-distribution and out-of-distribution optimization. Our method can automatically split in-distribution and out-of-distribution radiance fields.
\begin{equation} \label{eq:low_res_loss}
    \mathcal{L}^{LR} = \sum_{t=1}^{N} \mathcal{L}^{C}_{t} + \lambda_{\Delta}\mathcal{L}_{\Delta} + \lambda_{w} \mathcal{L}_{w} + (\lambda_{\mathcal{D}} \mathcal{L}_{\mathcal{D}}) \,,
\end{equation}

\subsection{Editing directions}
In our paper, we apply InterfaceGAN~\cite{shen2020interpreting} and StyleCLIP~\cite{patashnik2021styleclip} to EG3D.

\topic{InterfaceGAN.} InterfaceGAN requires a pre-trained classifier to label synthetic image data generated from the generator. We use CLIP~\cite{radford2021clip} instead to label data. For each image, we use the text prompt ``A portrait of a X face.'', where ``X'' is the attribute, \eg, smiling, old-man, and then compute the CLIP score. The CLIP scores act as the labels and we then apply InterfaceGAN to labelled image-score pairs for training.
For each editing direction, we generate 500,000 image-score pairs for training.

\topic{StyleCLIP.} We apply StyleCLIP mapper~\cite{patashnik2021styleclip} to synthetic images generated from pretrained EG3D generator. For each editing direction, we use 50,000 synthetic images for training, and 10,000 images for validation.

\begin{figure}[h!]
    \centering

    \begin{subfigure}[b]{.22\textwidth}
        \centering
        \frame{\includegraphics[width=\linewidth]{images/rebuttal/visualizations/frame0000.png}}
        \caption{\small{Input}}
        \label{subfig:input}
    \end{subfigure}\hfill
    \begin{subfigure}[b]{.22\textwidth}
        \centering
        \frame{\includegraphics[width=\linewidth]{images/rebuttal/visualizations/viz_mask_ood.png}}
        \caption{\small{Mask}}
        \label{subfig:mask}
    \end{subfigure}\hfill
    \begin{subfigure}[b]{.22\textwidth}
        \centering
        \frame{\includegraphics[width=\linewidth]{images/rebuttal/visualizations/viz_plane_xy_ood.png}}
        \caption{\small{XY plane}}
        \label{subfig:xyplane}
    \end{subfigure}\hfill
    \begin{subfigure}[b]{.22\textwidth}
        \centering
        \frame{\includegraphics[width=\linewidth]{images/rebuttal/visualizations/viz_plane_xz_ood.png}}
        \caption{\small{XZ plane}}
        \label{subfig:xzplane}
    \end{subfigure}

    \caption{
    \textbf{Visualization of learned mask and feature maps}. We visualize the learned mask and different tri-plane feature maps. The learned mask indicates the possibility of each pixel as an OOD object. 
    }
    \label{fig:common_video}
\end{figure}

\section{Additional results}
\subsection{Single Image Reconstruction, editing and view synthesis}
Please refer to our \href{https://in-n-out-3d.github.io/}{project page} for more results.

\subsection{Video Reconstruction, editing and view synthesis}
We show more results in our project page. Please refer to \href{https://in-n-out-3d.github.io/}{project page}.

\subsection{Non-face data}
Our method also applies to non-face data. 
Here we show an example of cat. 
We use the same pipeline but with a cat EG3D generator.
The editing direction is ``black cat'' from GOAE~\cite{Yuan_2023_GOAE}.
\begin{figure}[h!]
    \centering
    \begin{subfigure}[b]{0.3\linewidth}
        \centering
        \includegraphics[width=0.75\textwidth]{images/rebuttal/cat/frame0000_sunglasses.png}
        \caption{Input}
    \end{subfigure}
    \hfill % adds horizontal space between figures
    \begin{subfigure}[b]{0.3\linewidth}
        \centering
        \includegraphics[width=0.75\textwidth]{images/rebuttal/cat/proj_00000.png}
        \caption{Inv.}
    \end{subfigure}
    \hfill
    \begin{subfigure}[b]{0.3\linewidth}
        \centering
        \includegraphics[width=0.75\textwidth]{images/rebuttal/cat/proj_00000_black.png}
        \caption{``Black cat''}
    \end{subfigure}
    % \caption{Three images side by side}
    \vspace{-5mm}
\end{figure}

% \subsection{Video Editing}
% After the reconstruction, we can modify the latent code $w_t$ to achieve semantic editing. 
% Because we split the image into two radiance fields, out-of-distribution has little to do with semantic editing, during the editing, we only edit the latent code to get edited images. 
% At this stage, any existing GAN-based editing approaches can be used.
% Our experiments mainly study InterfaceGAN~\cite{shen2020interpreting} and StyleCLIP~\cite{patashnik2021styleclip}.
% Please refer to \href{run:./index.html}{index.html} for more visual results. 

% \subsection{Novel view synthesis}
% We can also render novel views after inversion with OOD objects. 
% Please refer to our Please refer to \href{run:./index.html}{index.html} for more visual results. 

\subsection{OOD object removal}
By setting the blending weights of the OOD objects to 0, we can remove OOD objects.
Please refer to \href{https://in-n-out-3d.github.io/}{project page} for more visual results. 

\subsection{Limitations/Failure cases}
We visualize two limitations in our \href{https://in-n-out-3d.github.io/}{project page}.

\topic{Floaters.} Our method introduces a new tri-plane, for single image inversion, its view synthesis result may involve some ``floaters'', since single-image 3D reconstruction is an ill-posed problem. Possible solution includes using distortion loss~\cite{barron2021mip}.
However, except for the novel views, our method outperforms other methods in terms of reconstruction quality and achieves faithful editing. 

\subsection{Visualization of learned masks and triplane features}
We show an example in Figure~\ref{fig:common_video}.
Please be aware of a \href{https://github.com/NVlabs/eg3d/issues/67}{bug} in the official EG3D tri-plane implementation, where the tri-planes are actually XY, XZ, and ZX. Here we only visualize XY and XZ planes.

%% file: algorithms/NN_initial.tex
\begin{algorithm}[h!]
\caption{NN\_init(): Nearest Neighbor (NN) Initialization for latent codes} \label{alg:nn_init}
\SetAlgoLined
\SetKwInOut{Input}{Input}
\SetKwInOut{Output}{Output}

\Input{
Image $\mathbf{I}_t$, or Video $\mathbf{V}=[\mathbf{I}_1, \cdots, \mathbf{I}_N]$, 
pre-trained generator $G$, 
camera parameters $[p_1, \cdots, p_N]$,
perceptual loss $LPIPS()$,
the number of samples $n$.
}
\Output{
Initialized latent codes $[w_1, \cdots, w_N]$.
}

\For{$t \gets 1$ to $N$}{
    $w_{samples} = []$ \\
    Distances $dists = []$ \\
    \For{$i \gets 1$ to $n$}{
        {Draw $z_i \in \mathbb{R}^{512} \sim \mathcal{N}(\mathbf{0}, \mathbf{I})$.} \\
        {Compute $w_i \gets G.mapping(z_i, p_t )$} \\
        {$w_{samples}$.insert($w_i$)} \\
        {$dists$.insert($LPIPS(G(w_i, p_t), \mathbf{I}_t)$)}
    }
    $w_t \gets NearestNeighbor(w_{samples}, dists)$    
}

\textbf{Return} mean($[w_1, \cdots, w_N]$).
 
\end{algorithm}

%% file: main.bbl
\begin{thebibliography}{71}
\providecommand{\natexlab}[1]{#1}
\providecommand{\url}[1]{\texttt{#1}}
\expandafter\ifx\csname urlstyle\endcsname\relax
  \providecommand{\doi}[1]{doi: #1}\else
  \providecommand{\doi}{doi: \begingroup \urlstyle{rm}\Url}\fi

\bibitem[Abdal et~al.(2019)Abdal, Qin, and Wonka]{abdal2019image2stylegan}
Rameen Abdal, Yipeng Qin, and Peter Wonka.
\newblock Image2stylegan: How to embed images into the stylegan latent space?
\newblock In \emph{ICCV}, 2019.

\bibitem[Abdal et~al.(2020)Abdal, Qin, and Wonka]{abdal2020image2stylegan++}
Rameen Abdal, Yipeng Qin, and Peter Wonka.
\newblock Image2stylegan++: How to edit the embedded images?
\newblock In \emph{CVPR}, 2020.

\bibitem[Abdal et~al.(2021)Abdal, Zhu, Mitra, and Wonka]{abdal2021styleflow}
Rameen Abdal, Peihao Zhu, Niloy~J Mitra, and Peter Wonka.
\newblock Styleflow: Attribute-conditioned exploration of stylegan-generated images using conditional continuous normalizing flows.
\newblock \emph{ACM Transactions on Graphics (ToG)}, 40\penalty0 (3):\penalty0 1--21, 2021.

\bibitem[Alaluf et~al.(2021)Alaluf, Patashnik, and Cohen-Or]{alaluf2021restyle}
Yuval Alaluf, Or Patashnik, and Daniel Cohen-Or.
\newblock Restyle: A residual-based stylegan encoder via iterative refinement.
\newblock In \emph{ICCV}, 2021.

\bibitem[Alaluf et~al.(2022)Alaluf, Tov, Mokady, Gal, and Bermano]{alaluf2022hyperstyle}
Yuval Alaluf, Omer Tov, Ron Mokady, Rinon Gal, and Amit Bermano.
\newblock Hyperstyle: Stylegan inversion with hypernetworks for real image editing.
\newblock In \emph{CVPR}, 2022.

\bibitem[Barron et~al.(2021)Barron, Mildenhall, Tancik, Hedman, Martin-Brualla, and Srinivasan]{barron2021mip}
Jonathan~T Barron, Ben Mildenhall, Matthew Tancik, Peter Hedman, Ricardo Martin-Brualla, and Pratul~P Srinivasan.
\newblock Mip-nerf: A multiscale representation for anti-aliasing neural radiance fields.
\newblock In \emph{ICCV}, 2021.

\bibitem[Bau et~al.(2020)Bau, Strobelt, Peebles, Zhou, Zhu, Torralba, et~al.]{bau2020semantic}
David Bau, Hendrik Strobelt, William Peebles, Bolei Zhou, Jun-Yan Zhu, Antonio Torralba, et~al.
\newblock Semantic photo manipulation with a generative image prior.
\newblock \emph{ACM Transactions on Graphics (ToG)}, 38\penalty0 (4):\penalty0 1--11, 2020.

\bibitem[Chai et~al.(2021)Chai, Wulff, and Isola]{chai2021latent}
Lucy Chai, Jonas Wulff, and Phillip Isola.
\newblock Using latent space regression to analyze and leverage compositionality in gans.
\newblock In \emph{ICLR}, 2021.

\bibitem[Chan et~al.(2021)Chan, Monteiro, Kellnhofer, Wu, and Wetzstein]{chan2021pi}
Eric~R Chan, Marco Monteiro, Petr Kellnhofer, Jiajun Wu, and Gordon Wetzstein.
\newblock pi-gan: Periodic implicit generative adversarial networks for 3d-aware image synthesis.
\newblock In \emph{CVPR}, 2021.

\bibitem[Chan et~al.(2022)Chan, Lin, Chan, Nagano, Pan, De~Mello, Gallo, Guibas, Tremblay, Khamis, et~al.]{chan2022efficient}
Eric~R Chan, Connor~Z Lin, Matthew~A Chan, Koki Nagano, Boxiao Pan, Shalini De~Mello, Orazio Gallo, Leonidas~J Guibas, Jonathan Tremblay, Sameh Khamis, et~al.
\newblock Efficient geometry-aware 3d generative adversarial networks.
\newblock In \emph{CVPR}, 2022.

\bibitem[Chen et~al.(2022)Chen, Xu, Geiger, Yu, and Su]{Chen2022ECCV}
Anpei Chen, Zexiang Xu, Andreas Geiger, Jingyi Yu, and Hao Su.
\newblock Tensorf: Tensorial radiance fields.
\newblock In \emph{ECCV}, 2022.

\bibitem[Collins et~al.(2020)Collins, Bala, Price, and Susstrunk]{collins2020editing}
Edo Collins, Raja Bala, Bob Price, and Sabine Susstrunk.
\newblock Editing in style: Uncovering the local semantics of gans.
\newblock In \emph{CVPR}, 2020.

\bibitem[Daras et~al.(2020)Daras, Odena, Zhang, and Dimakis]{daras2020your}
Giannis Daras, Augustus Odena, Han Zhang, and Alexandros~G Dimakis.
\newblock Your local gan: Designing two dimensional local attention mechanisms for generative models.
\newblock In \emph{CVPR}, 2020.

\bibitem[Deng et~al.(2019{\natexlab{a}})Deng, Guo, Xue, and Zafeiriou]{deng2019arcface}
Jiankang Deng, Jia Guo, Niannan Xue, and Stefanos Zafeiriou.
\newblock Arcface: Additive angular margin loss for deep face recognition.
\newblock In \emph{CVPR}, 2019{\natexlab{a}}.

\bibitem[Deng et~al.(2019{\natexlab{b}})Deng, Yang, Xu, Chen, Jia, and Tong]{deng2019accurate}
Yu Deng, Jiaolong Yang, Sicheng Xu, Dong Chen, Yunde Jia, and Xin Tong.
\newblock Accurate 3d face reconstruction with weakly-supervised learning: From single image to image set.
\newblock In \emph{CVPR Workshops}, 2019{\natexlab{b}}.

\bibitem[Deng et~al.(2022)Deng, Yang, Xiang, and Tong]{deng2022gram}
Yu Deng, Jiaolong Yang, Jianfeng Xiang, and Xin Tong.
\newblock Gram: Generative radiance manifolds for 3d-aware image generation.
\newblock In \emph{CVPR}, 2022.

\bibitem[Fr{\"u}hst{\"u}ck et~al.(2023)Fr{\"u}hst{\"u}ck, Sarafianos, Xu, Wonka, and Tung]{fruhstuck2023vive3d}
Anna Fr{\"u}hst{\"u}ck, Nikolaos Sarafianos, Yuanlu Xu, Peter Wonka, and Tony Tung.
\newblock Vive3d: Viewpoint-independent video editing using 3d-aware gans.
\newblock In \emph{CVPR}, 2023.

\bibitem[Gal et~al.(2022)Gal, Patashnik, Maron, Bermano, Chechik, and Cohen-Or]{gal2022stylegan}
Rinon Gal, Or Patashnik, Haggai Maron, Amit~H Bermano, Gal Chechik, and Daniel Cohen-Or.
\newblock Stylegan-nada: Clip-guided domain adaptation of image generators.
\newblock \emph{ACM Transactions on Graphics (TOG)}, 41\penalty0 (4):\penalty0 1--13, 2022.

\bibitem[Gao et~al.(2021)Gao, Saraf, Kopf, and Huang]{gao2021dynamic}
Chen Gao, Ayush Saraf, Johannes Kopf, and Jia-Bin Huang.
\newblock Dynamic view synthesis from dynamic monocular video.
\newblock In \emph{ICCV}, 2021.

\bibitem[Gao et~al.(2022)Gao, Shen, Wang, Chen, Yin, Li, Litany, Gojcic, and Fidler]{gao2022get3d}
Jun Gao, Tianchang Shen, Zian Wang, Wenzheng Chen, Kangxue Yin, Daiqing Li, Or Litany, Zan Gojcic, and Sanja Fidler.
\newblock Get3d: A generative model of high quality 3d textured shapes learned from images.
\newblock \emph{arXiv preprint arXiv:2209.11163}, 2022.

\bibitem[Gu et~al.(2020)Gu, Shen, and Zhou]{gu2020image}
Jinjin Gu, Yujun Shen, and Bolei Zhou.
\newblock Image processing using multi-code gan prior.
\newblock In \emph{CVPR}, 2020.

\bibitem[Gu et~al.(2022)Gu, Liu, Wang, and Theobalt]{gu2022stylenerf}
Jiatao Gu, Lingjie Liu, Peng Wang, and Christian Theobalt.
\newblock Stylene{RF}: A style-based 3d aware generator for high-resolution image synthesis.
\newblock In \emph{ICLR}, 2022.

\bibitem[Guo et~al.(2020)Guo, Zhu, Yang, Yang, Lei, and Li]{guo2020towards}
Jianzhu Guo, Xiangyu Zhu, Yang Yang, Fan Yang, Zhen Lei, and Stan~Z Li.
\newblock Towards fast, accurate and stable 3d dense face alignment.
\newblock In \emph{ECCV}, 2020.

\bibitem[H{\"a}rk{\"o}nen et~al.(2020)H{\"a}rk{\"o}nen, Hertzmann, Lehtinen, and Paris]{harkonen2020ganspace}
Erik H{\"a}rk{\"o}nen, Aaron Hertzmann, Jaakko Lehtinen, and Sylvain Paris.
\newblock Ganspace: Discovering interpretable gan controls.
\newblock In \emph{NeurIPS}, 2020.

\bibitem[Huh et~al.(2020)Huh, Zhang, Zhu, Paris, and Hertzmann]{huh2020ganprojection}
Minyoung Huh, Richard Zhang, Jun-Yan Zhu, Sylvain Paris, and Aaron Hertzmann.
\newblock Transforming and projecting images to class-conditional generative networks.
\newblock In \emph{ECCV}, 2020.

\bibitem[Karras et~al.(2019)Karras, Laine, and Aila]{karras2019style}
Tero Karras, Samuli Laine, and Timo Aila.
\newblock A style-based generator architecture for generative adversarial networks.
\newblock In \emph{CVPR}, 2019.

\bibitem[Karras et~al.(2020{\natexlab{a}})Karras, Aittala, Hellsten, Laine, Lehtinen, and Aila]{karras2020training}
Tero Karras, Miika Aittala, Janne Hellsten, Samuli Laine, Jaakko Lehtinen, and Timo Aila.
\newblock Training generative adversarial networks with limited data.
\newblock In \emph{NeurIPS}, 2020{\natexlab{a}}.

\bibitem[Karras et~al.(2020{\natexlab{b}})Karras, Laine, Aittala, Hellsten, Lehtinen, and Aila]{karras2020analyzing}
Tero Karras, Samuli Laine, Miika Aittala, Janne Hellsten, Jaakko Lehtinen, and Timo Aila.
\newblock Analyzing and improving the image quality of stylegan.
\newblock In \emph{CVPR}, 2020{\natexlab{b}}.

\bibitem[Karras et~al.(2021)Karras, Aittala, Laine, H{\"a}rk{\"o}nen, Hellsten, Lehtinen, and Aila]{karras2021alias}
Tero Karras, Miika Aittala, Samuli Laine, Erik H{\"a}rk{\"o}nen, Janne Hellsten, Jaakko Lehtinen, and Timo Aila.
\newblock Alias-free generative adversarial networks.
\newblock In \emph{NeurIPS}, 2021.

\bibitem[Kingma and Ba(2014)]{kingma2014adam}
Diederik~P Kingma and Jimmy Ba.
\newblock Adam: A method for stochastic optimization.
\newblock \emph{arXiv preprint arXiv:1412.6980}, 2014.

\bibitem[Lan et~al.(2023)Lan, Meng, Yang, Loy, and Dai]{lan2023e3dge}
Yushi Lan, Xuyi Meng, Shuai Yang, Chen~Change Loy, and Bo Dai.
\newblock Self-supervised geometry-aware encoder for style-based 3d gan inversion.
\newblock In \emph{CVPR}, 2023.

\bibitem[Lin et~al.(2022)Lin, Lindell, Chan, and Wetzstein]{lin20223dganinversion}
C.Z. Lin, D.B. Lindell, E.R. Chan, and G. Wetzstein.
\newblock 3d gan inversion for controllable portrait image animation.
\newblock In \emph{ECCVW}, 2022.

\bibitem[Luo et~al.(2017)Luo, Xu, Tang, and Lv]{luo2017learning}
Junyu Luo, Yong Xu, Chenwei Tang, and Jiancheng Lv.
\newblock Learning inverse mapping by autoencoder based generative adversarial nets.
\newblock In \emph{NeurIPS}, 2017.

\bibitem[Martin-Brualla et~al.(2021)Martin-Brualla, Radwan, Sajjadi, Barron, Dosovitskiy, and Duckworth]{martin2021nerf}
Ricardo Martin-Brualla, Noha Radwan, Mehdi~SM Sajjadi, Jonathan~T Barron, Alexey Dosovitskiy, and Daniel Duckworth.
\newblock Nerf in the wild: Neural radiance fields for unconstrained photo collections.
\newblock In \emph{CVPR}, 2021.

\bibitem[Max(1995)]{max1995optical}
Nelson Max.
\newblock Optical models for direct volume rendering.
\newblock \emph{IEEE Transactions on Visualization and Computer Graphics}, 1\penalty0 (2):\penalty0 99--108, 1995.

\bibitem[Mildenhall et~al.(2020)Mildenhall, Srinivasan, Tancik, Barron, Ramamoorthi, and Ng]{mildenhall2020nerf}
Ben Mildenhall, Pratul~P. Srinivasan, Matthew Tancik, Jonathan~T. Barron, Ravi Ramamoorthi, and Ren Ng.
\newblock Nerf: Representing scenes as neural radiance fields for view synthesis.
\newblock In \emph{ECCV}, 2020.

\bibitem[M\"uller et~al.(2022)M\"uller, Evans, Schied, and Keller]{mueller2022instant}
Thomas M\"uller, Alex Evans, Christoph Schied, and Alexander Keller.
\newblock Instant neural graphics primitives with a multiresolution hash encoding.
\newblock \emph{ACM Trans. Graph.}, 41\penalty0 (4):\penalty0 102:1--102:15, 2022.

\bibitem[Nguyen-Phuoc et~al.(2019)Nguyen-Phuoc, Li, Theis, Richardt, and Yang]{nguyen2019hologan}
Thu Nguyen-Phuoc, Chuan Li, Lucas Theis, Christian Richardt, and Yong-Liang Yang.
\newblock Hologan: Unsupervised learning of 3d representations from natural images.
\newblock In \emph{ICCV}, 2019.

\bibitem[Nitzan et~al.(2020)Nitzan, Bermano, Li, and Cohen-Or]{Nitzan2020FaceID}
Yotam Nitzan, A. Bermano, Yangyan Li, and D. Cohen-Or.
\newblock Face identity disentanglement via latent space mapping.
\newblock \emph{ACM Transactions on Graphics (TOG)}, 39:\penalty0 1 -- 14, 2020.

\bibitem[Or-El et~al.(2022)Or-El, Luo, Shan, Shechtman, Park, and Kemelmacher-Shlizerman]{orel2022styleSDF}
Roy Or-El, Xuan Luo, Mengyi Shan, Eli Shechtman, Jeong~Joon Park, and Ira Kemelmacher-Shlizerman.
\newblock Style{SDF}: {H}igh-{R}esolution {3D}-{C}onsistent {I}mage and {G}eometry {G}eneration.
\newblock In \emph{CVPR}, 2022.

\bibitem[Patashnik et~al.(2021)Patashnik, Wu, Shechtman, Cohen-Or, and Lischinski]{patashnik2021styleclip}
Or Patashnik, Zongze Wu, Eli Shechtman, Daniel Cohen-Or, and Dani Lischinski.
\newblock Styleclip: Text-driven manipulation of stylegan imagery.
\newblock In \emph{ICCV}, 2021.

\bibitem[Raj et~al.(2019)Raj, Li, and Bresler]{raj2019gan}
Ankit Raj, Yuqi Li, and Yoram Bresler.
\newblock Gan-based projector for faster recovery with convergence guarantees in linear inverse problems.
\newblock In \emph{ICCV}, 2019.

\bibitem[Ranftl et~al.(2022)Ranftl, Lasinger, Hafner, Schindler, and Koltun]{Ranftl2022midas}
Ren\'{e} Ranftl, Katrin Lasinger, David Hafner, Konrad Schindler, and Vladlen Koltun.
\newblock Towards robust monocular depth estimation: Mixing datasets for zero-shot cross-dataset transfer.
\newblock \emph{TPAMI}, 44\penalty0 (3), 2022.

\bibitem[Richardson et~al.(2021)Richardson, Alaluf, Patashnik, Nitzan, Azar, Shapiro, and Cohen-Or]{richardson2021encoding}
Elad Richardson, Yuval Alaluf, Or Patashnik, Yotam Nitzan, Yaniv Azar, Stav Shapiro, and Daniel Cohen-Or.
\newblock Encoding in style: a stylegan encoder for image-to-image translation.
\newblock In \emph{CVPR}, 2021.

\bibitem[Roich et~al.(2022)Roich, Mokady, Bermano, and Cohen-Or]{roich2022pivotal}
Daniel Roich, Ron Mokady, Amit~H Bermano, and Daniel Cohen-Or.
\newblock Pivotal tuning for latent-based editing of real images.
\newblock \emph{ACM Transactions on Graphics (TOG)}, 42\penalty0 (1):\penalty0 1--13, 2022.

\bibitem[Schwarz et~al.(2020)Schwarz, Liao, Niemeyer, and Geiger]{schwarz2020graf}
Katja Schwarz, Yiyi Liao, Michael Niemeyer, and Andreas Geiger.
\newblock Graf: Generative radiance fields for 3d-aware image synthesis.
\newblock In \emph{NeurIPS}, 2020.

\bibitem[Shen et~al.(2020)Shen, Gu, Tang, and Zhou]{shen2020interpreting}
Yujun Shen, Jinjin Gu, Xiaoou Tang, and Bolei Zhou.
\newblock Interpreting the latent space of gans for semantic face editing.
\newblock In \emph{CVPR}, 2020.

\bibitem[Simsar et~al.(2023)Simsar, Tonioni, Ornek, and Tombari]{Simsar_2023_swap3d}
Enis Simsar, Alessio Tonioni, Evin~Pinar Ornek, and Federico Tombari.
\newblock Latentswap3d: Semantic edits on 3d image gans.
\newblock In \emph{ICCVW}, 2023.

\bibitem[Skorokhodov et~al.(2022)Skorokhodov, Tulyakov, Wang, and Wonka]{skorokhodov2022epigraf}
Ivan Skorokhodov, Sergey Tulyakov, Yiqun Wang, and Peter Wonka.
\newblock Epigraf: Rethinking training of 3d gans.
\newblock \emph{arXiv preprint arXiv:2206.10535}, 2022.

\bibitem[{\v{S}}ubrtov{\'a} et~al.(2022){\v{S}}ubrtov{\'a}, Futschik, {\v{C}}ech, Luk{\'a}{\v{c}}, Shechtman, and S{\`y}kora]{vsubrtova2022chunkygan}
Ad{\'e}la {\v{S}}ubrtov{\'a}, David Futschik, Jan {\v{C}}ech, Michal Luk{\'a}{\v{c}}, Eli Shechtman, and Daniel S{\`y}kora.
\newblock Chunkygan: Real image inversion via segments.
\newblock In \emph{European Conference on Computer Vision}, 2022.

\bibitem[Sun et~al.(2022)Sun, Wang, Shi, Wang, Wang, and Liu]{sun2022ide}
Jingxiang Sun, Xuan Wang, Yichun Shi, Lizhen Wang, Jue Wang, and Yebin Liu.
\newblock Ide-3d: Interactive disentangled editing for high-resolution 3d-aware portrait synthesis.
\newblock \emph{ACM Transactions on Graphics (ToG)}, 41\penalty0 (6):\penalty0 1--10, 2022.

\bibitem[Tancik et~al.(2020)Tancik, Srinivasan, Mildenhall, Fridovich-Keil, Raghavan, Singhal, Ramamoorthi, Barron, and Ng]{tancik2020fourier}
Matthew Tancik, Pratul Srinivasan, Ben Mildenhall, Sara Fridovich-Keil, Nithin Raghavan, Utkarsh Singhal, Ravi Ramamoorthi, Jonathan Barron, and Ren Ng.
\newblock Fourier features let networks learn high frequency functions in low dimensional domains.
\newblock In \emph{NeurIPS}, 2020.

\bibitem[Tewari et~al.(2020{\natexlab{a}})Tewari, Elgharib, Bernard, Seidel, P{\'e}rez, Zollh{\"o}fer, Theobalt, et~al.]{tewari2020pie}
Ayush Tewari, Mohamed Elgharib, Florian Bernard, Hans-Peter Seidel, Patrick P{\'e}rez, Michael Zollh{\"o}fer, Christian Theobalt, et~al.
\newblock Pie: Portrait image embedding for semantic control.
\newblock \emph{arXiv preprint arXiv:2009.09485}, 2020{\natexlab{a}}.

\bibitem[Tewari et~al.(2020{\natexlab{b}})Tewari, Elgharib, Bharaj, Bernard, Seidel, P{\'e}rez, Zollhofer, and Theobalt]{tewari2020stylerig}
Ayush Tewari, Mohamed Elgharib, Gaurav Bharaj, Florian Bernard, Hans-Peter Seidel, Patrick P{\'e}rez, Michael Zollhofer, and Christian Theobalt.
\newblock Stylerig: Rigging stylegan for 3d control over portrait images.
\newblock In \emph{CVPR}, 2020{\natexlab{b}}.

\bibitem[Tov et~al.(2021)Tov, Alaluf, Nitzan, Patashnik, and Cohen-Or]{tov2021designing}
Omer Tov, Yuval Alaluf, Yotam Nitzan, Or Patashnik, and Daniel Cohen-Or.
\newblock Designing an encoder for stylegan image manipulation.
\newblock \emph{ACM Transactions on Graphics (TOG)}, 40\penalty0 (4):\penalty0 1--14, 2021.

\bibitem[Trevithick et~al.(2023)Trevithick, Chan, Stengel, Chan, Liu, Yu, Khamis, Chandraker, Ramamoorthi, and Nagano]{trevithick2023real}
Alex Trevithick, Matthew Chan, Michael Stengel, Eric Chan, Chao Liu, Zhiding Yu, Sameh Khamis, Manmohan Chandraker, Ravi Ramamoorthi, and Koki Nagano.
\newblock Real-time radiance fields for single-image portrait view synthesis.
\newblock \emph{ACM Transactions on Graphics (TOG)}, 42\penalty0 (4):\penalty0 1--15, 2023.

\bibitem[Tzaban et~al.(2022)Tzaban, Mokady, Gal, Bermano, and Cohen-Or]{tzaban2022stitch}
Rotem Tzaban, Ron Mokady, Rinon Gal, Amit~H Bermano, and Daniel Cohen-Or.
\newblock Stitch it in time: Gan-based facial editing of real videos.
\newblock \emph{SIGGRAPH Asia 2022 Conference Papers}, 2022.

\bibitem[Viazovetskyi et~al.(2020)Viazovetskyi, Ivashkin, and Kashin]{viazovetskyi2020stylegan2}
Yuri Viazovetskyi, Vladimir Ivashkin, and Evgeny Kashin.
\newblock Stylegan2 distillation for feed-forward image manipulation.
\newblock In \emph{ECCV}, 2020.

\bibitem[Wu et~al.(2022)Wu, Zhong, Tagliasacchi, Cole, and Oztireli]{wu2022d}
Tianhao Wu, Fangcheng Zhong, Andrea Tagliasacchi, Forrester Cole, and Cengiz Oztireli.
\newblock D$^2$nerf: Self-supervised decoupling of dynamic and static objects from a monocular video.
\newblock \emph{arXiv preprint arXiv:2205.15838}, 2022.

\bibitem[Wu et~al.(2021)Wu, Lischinski, and Shechtman]{wu2021stylespace}
Zongze Wu, Dani Lischinski, and Eli Shechtman.
\newblock Stylespace analysis: Disentangled controls for stylegan image generation.
\newblock In \emph{CVPR}, 2021.

\bibitem[Xia et~al.(2022)Xia, Zhang, Yang, Xue, Zhou, and Yang]{xia2022gan}
Weihao Xia, Yulun Zhang, Yujiu Yang, Jing-Hao Xue, Bolei Zhou, and Ming-Hsuan Yang.
\newblock Gan inversion: A survey.
\newblock \emph{IEEE Transactions on Pattern Analysis and Machine Intelligence}, 2022.

\bibitem[Xie et~al.(2023)Xie, Ouyang, Piao, Lei, and Chen]{xie2023hfgi3d}
Jiaxin Xie, Hao Ouyang, Jingtan Piao, Chenyang Lei, and Qifeng Chen.
\newblock High-fidelity 3d gan inversion by pseudo-multi-view optimization.
\newblock In \emph{CVPR}, 2023.

\bibitem[Xu et~al.(2022)Xu, AlBahar, and Huang]{xu2022temporally}
Yiran Xu, Badour AlBahar, and Jia-Bin Huang.
\newblock Temporally consistent semantic video editing.
\newblock In \emph{ECCV}, 2022.

\bibitem[Yang et~al.(2021)Yang, Zhang, Xu, Li, Zhou, Bao, Zhang, and Cui]{yang2021learning}
Bangbang Yang, Yinda Zhang, Yinghao Xu, Yijin Li, Han Zhou, Hujun Bao, Guofeng Zhang, and Zhaopeng Cui.
\newblock Learning object-compositional neural radiance field for editable scene rendering.
\newblock In \emph{ICCV}, 2021.

\bibitem[Yao et~al.(2021)Yao, Newson, Gousseau, and Hellier]{yao2021latent}
Xu Yao, Alasdair Newson, Yann Gousseau, and Pierre Hellier.
\newblock A latent transformer for disentangled face editing in images and videos.
\newblock In \emph{ICCV}, 2021.

\bibitem[Yin et~al.(2023)Yin, Zhang, Wang, Wang, Li, Gong, Fan, Cun, Shan, Oztireli, et~al.]{yin2023spi}
Fei Yin, Yong Zhang, Xuan Wang, Tengfei Wang, Xiaoyu Li, Yuan Gong, Yanbo Fan, Xiaodong Cun, Ying Shan, Cengiz Oztireli, et~al.
\newblock 3d gan inversion with facial symmetry prior.
\newblock In \emph{CVPR}, 2023.

\bibitem[Yuan et~al.(2023)Yuan, Zhu, Li, Liu, and Yuan]{Yuan_2023_GOAE}
Ziyang Yuan, Yiming Zhu, Yu Li, Hongyu Liu, and Chun Yuan.
\newblock Make encoder great again in 3d gan inversion through geometry and occlusion-aware encoding.
\newblock In \emph{ICCV}, 2023.

\bibitem[Zhang et~al.(2022)Zhang, Siarohin, Liu, Tang, Sebe, and Wang]{zhang2022training}
Jichao Zhang, Aliaksandr Siarohin, Yahui Liu, Hao Tang, Nicu Sebe, and Wei Wang.
\newblock Training and tuning generative neural radiance fields for attribute-conditional 3d-aware face generation.
\newblock \emph{arXiv preprint arXiv:2208.12550}, 2022.

\bibitem[Zhang et~al.(2018)Zhang, Isola, Efros, Shechtman, and Wang]{zhang2018perceptual}
Richard Zhang, Phillip Isola, Alexei~A Efros, Eli Shechtman, and Oliver Wang.
\newblock The unreasonable effectiveness of deep features as a perceptual metric.
\newblock In \emph{CVPR}, 2018.

\bibitem[Zhu et~al.(2020{\natexlab{a}})Zhu, Shen, Zhao, and Zhou]{zhu2020domain}
Jiapeng Zhu, Yujun Shen, Deli Zhao, and Bolei Zhou.
\newblock In-domain gan inversion for real image editing.
\newblock In \emph{European conference on computer vision}, 2020{\natexlab{a}}.

\bibitem[Zhu et~al.(2020{\natexlab{b}})Zhu, Shen, Zhao, and Zhou]{zhu2020indomain}
Jiapeng Zhu, Yujun Shen, Deli Zhao, and Bolei Zhou.
\newblock In-domain gan inversion for real image editing.
\newblock In \emph{ECCV}, 2020{\natexlab{b}}.

\end{thebibliography}
